\documentclass[letterpaper]{article} 
\usepackage{aaai25}  
\usepackage{times}  
\usepackage{helvet}  
\usepackage{courier}  
\usepackage[hyphens]{url}  
\usepackage{graphicx} 
\usepackage{natbib}  
\usepackage{caption} 
\usepackage{amsmath}
\usepackage{amssymb}
\usepackage{graphicx}
\usepackage{array}
\usepackage{caption}
\usepackage{algorithm}
\usepackage{algpseudocode}
\usepackage{subcaption}
\newlength{\firstColWidth}
\usepackage{newfloat}
\usepackage{listings}
\DeclareCaptionStyle{ruled}{labelfont=normalfont,labelsep=colon,strut=off} 
\floatstyle{ruled}
\newfloat{listing}{tb}{lst}{}
\floatname{listing}{Listing}
\title{Accelerating Manufacturing Scale-Up from Material Discovery Using Agentic Web Navigation and Retrieval-Augmented AI for Process Engineering Schematics Design}
\author{
    Sakhinana Sagar Srinivas\textsuperscript{}\thanks{Equal contribution}\textsuperscript{}\thanks{Corresponding author},
    Akash Das\textsuperscript{*}, 
    Shivam Gupta, 
    Venkataramana Runkana\\
}
\affiliations{
    Tata Research  \\%
    \texttt{\{sagar.sakhinana, akash.das4, g.shivam4, venkat.runkana\}}@tcs.com\\
}

\usepackage{bibentry}

\begin{document}

\maketitle


\begin{abstract}
Process Flow Diagrams (PFDs) and Process and Instrumentation Diagrams (PIDs) are critical tools for industrial process design, control, and safety. However, the generation of precise and regulation-compliant diagrams remains a significant challenge, particularly in scaling breakthroughs from material discovery to industrial production in an era of automation and digitalization. This paper introduces an autonomous agentic framework to address these challenges through a two-stage approach involving knowledge acquisition and generation. The framework integrates specialized sub-agents for retrieving and synthesizing multimodal data from publicly available online sources and constructs ontological knowledge graphs using a Graph Retrieval-Augmented Generation (Graph RAG) paradigm. These capabilities enable the automation of diagram generation and open-domain question answering (ODQA) tasks with high contextual accuracy. Extensive empirical experiments demonstrate the framework's ability to deliver regulation-compliant diagrams with minimal expert intervention, highlighting its practical utility for industrial applications.
\vspace{-4mm}
\end{abstract}

\section{Introduction}
\begin{figure}[!ht]
    \centering
    \resizebox{1.0\linewidth}{!}{
        \hspace*{0mm}\includegraphics[keepaspectratio,trim=0.0cm 3cm 0cm 0.025cm,clip]{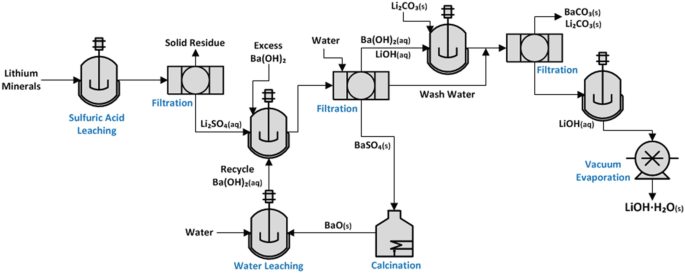} 
    }
    \vspace{-2mm}
    \caption{The figure shows the PFD for the production of lithium hydroxide (LiOH) from lithium minerals \cite{halim2022potential}. Lithium hydroxide is primarily used in lithium-ion battery production and high-purity chemical processes critical to semiconductor and electronics manufacturing, with a global market valued at \$10 billion in 2024 and projected to grow significantly.}
    \label{fig:figure2}
    \vspace{-3mm}
\end{figure}

\begin{figure}[!ht]
    \centering
    \resizebox{0.975\linewidth}{!}{
        \hspace*{0mm}\includegraphics[keepaspectratio,trim=0.0cm 0.025cm 0cm 0.025cm,clip]{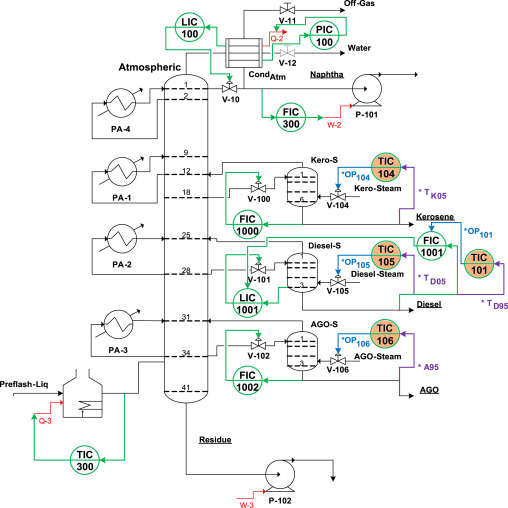} 
    }
    \vspace{-2mm}
    \caption{The figure shows the PID for the operation, maintenance, and troubleshooting of the atmospheric distillation unit \cite{sotelo2017design}, which separates crude oil into gasoline, naphtha, diesel, and gas oil—essential for fuel production. The global refining market was valued at approximately \$2 trillion in 2024, with projections to grow further.}
    \label{fig:figure3}
    \vspace{-3mm}
\end{figure}

\begin{figure}[!ht]
\vspace{-2mm}
\centering
\resizebox{1.0\linewidth}{!}{ 
\hspace*{0mm}\includegraphics[keepaspectratio,trim=0.0cm 0cm 0cm 0.025cm,clip]{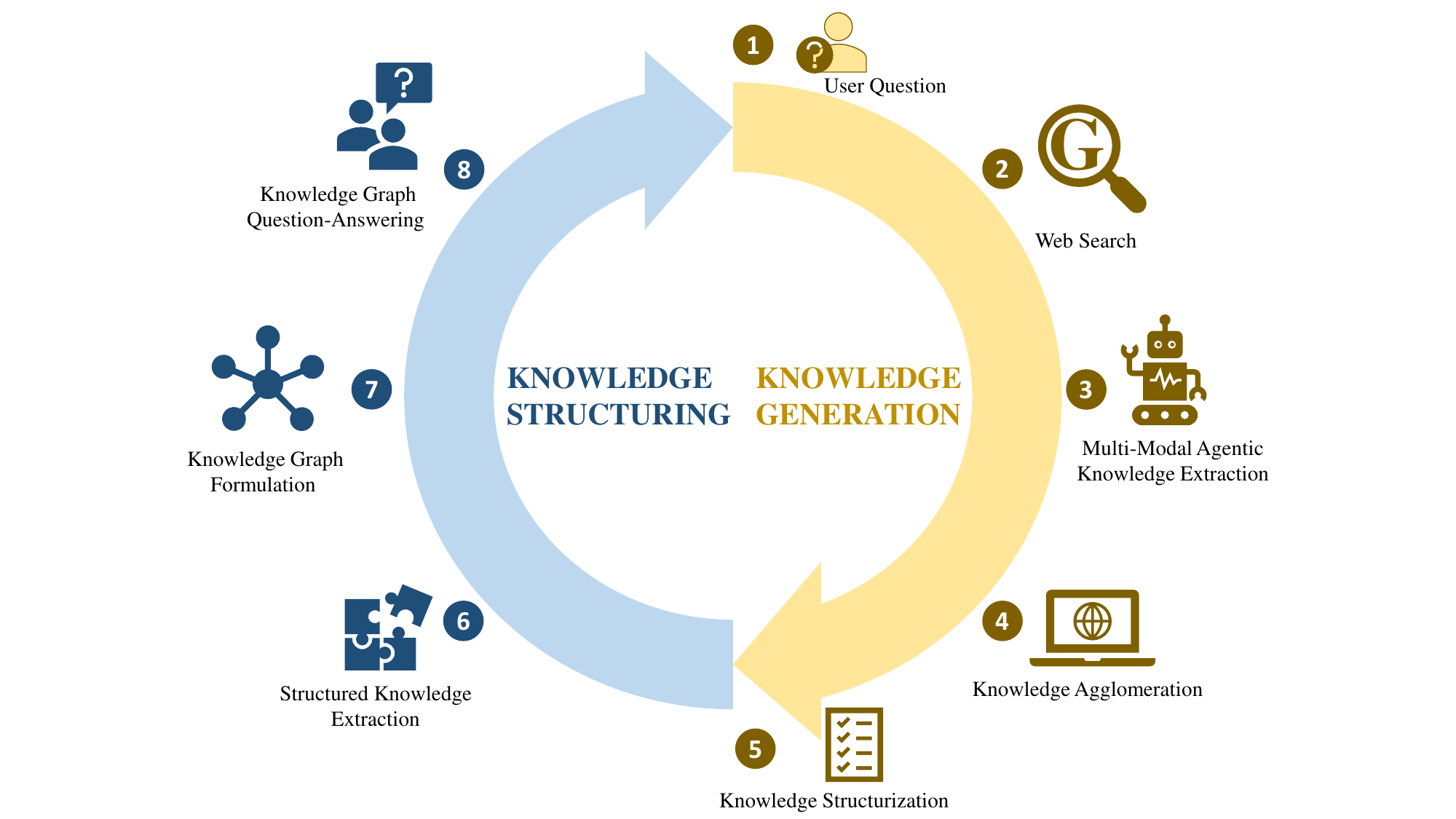} 
}
\vspace{-6mm}
\caption{We present a two-step approach to extract, aggregate, and structure knowledge for PFD and PID generation. In the first step, the autonomous agentic web navigation framework retrieves and synthesizes information from multiple online sources, enabling knowledge generation through chemical-specific web data aggregation. The second step involves two sub-steps: (a) extracting entity-relationship triples from the generated knowledge and populating an ontological knowledge graph, and (b) utilizing a Graph RAG framework to enable structured knowledge graph traversal and retrieval for complex ODQA tasks.}
\label{fig:figure4}
\vspace{-3mm}
\end{figure}

\begin{figure*}[!ht]
\vspace{-1mm}
\centering
\resizebox{0.925\linewidth}{!}{ 
\hspace*{0mm}\includegraphics[keepaspectratio,trim=0.0cm 0cm 0cm 0.10cm,clip]{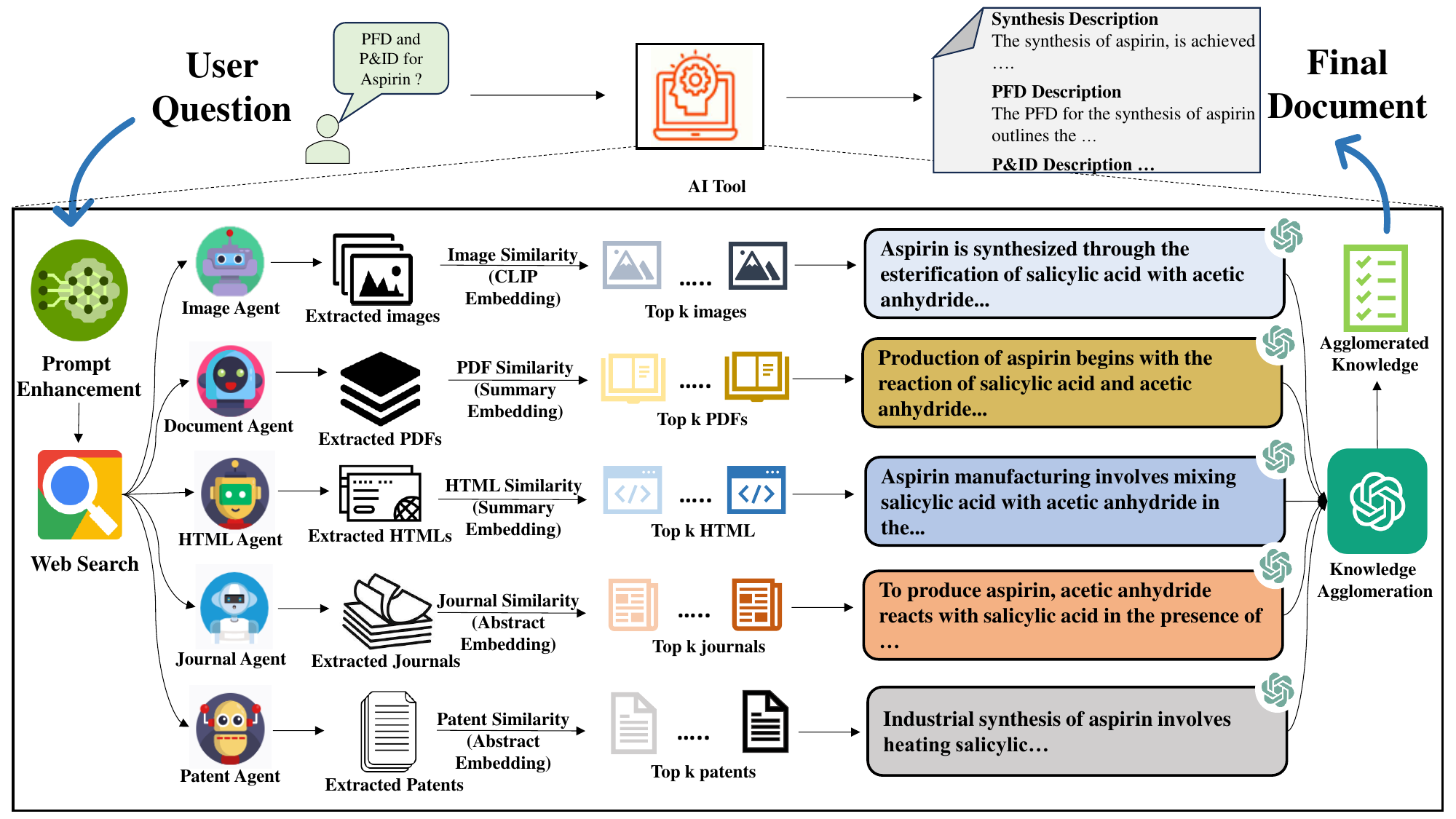} 
}
\vspace{-3mm}
\caption{The LLM-based multi-agent framework for knowledge generation retrieves and synthesizes information from diverse online sources on PFDs and PIDs for well-known chemicals in large-scale industrial synthesis. A meta-agent orchestrates specialized sub-agents that retrieve and analyze data sources such as images, scholarly articles, patents, and Wikipedia. Each sub-agent, dedicated to a specific retrieval task, uses tools like SerpAPI for accessing search results and LLMs (e.g., OpenAI GPT-4o) for synthesizing information. An iterative feedback mechanism involving human experts and advanced Gold language models ensures the accuracy and quality of the generated knowledge. This structured, feedback-driven approach optimizes knowledge generation in specialized domains like chemical process design and control. Ultimately, the framework constructs an ontological knowledge graph within the Graph RAG framework, enhancing context-aware retrieval for both complex ODQA tasks related to chemical process diagrams and the autogeneration of PFDs and PIDs for novel chemicals.}
\label{fig:figure5}
\vspace{-5mm}
\end{figure*}

PFDs (Process Flow Diagrams) and PIDs (Process and Instrumentation Diagrams) serve as the architectural blueprints for modern industrial operations (see Figures \ref{fig:figure2}, and \ref{fig:figure3}). They form the foundation of manufacturing by providing essential visualizations and operational details for process design and control, enabling efficient production, optimized industrial operations, and streamlined maintenance. These engineering tools underpin industrial design and operational control systems and are used in industries such as oil and gas, electronics manufacturing, pharmaceuticals, mining, mineral processing, and automotive and aerospace. PFDs offer a simplified macro-level overview of the entire process, illustrating material flows, energy balances, and equipment layouts, which facilitate initial design and optimization. In contrast, PIDs build upon PFDs and provide a micro-level perspective by detailing instrumentation and control schemes, which are essential for regulatory compliance, quality control, and safety in plant operation. In short, PFDs provide the context, while PIDs provide the specifics, making both indispensable throughout the entire lifecycle of a process plant, from initial design to decommissioning. As industries increasingly embrace automation and digitalization, the demand for accurate and efficient PFDs and PIDs continues to grow, highlighting their critical role in ensuring efficient, and compliant operations across various industrial sectors. Recent breakthroughs in generative AI are revolutionizing materials science and engineering\cite{jia2024llmatdesign, liu2023materials, kim2024materials, ansari2024dziner}, enabling autonomous material discovery and replacing expensive, time-consuming trial-and-error experimentation. However, these advancements often struggle to transition from computer simulations and lab experiments to large-scale industrial production. Scaling autonomous material discoveries to large-scale industrial production demands effective process design and control. PFDs and PIDs are crucial tools—providing high-level process overviews (material and energy balances) and detailed instrumentation/control strategies (consistent quality and safety) respectively—fundamental to the design, operation, and optimization of these processes. Automating their generation enables scalable solutions, transforming theoretical advances into economically viable industrial applications and accelerating the transition from research to real-world deployment, with transformative impacts across numerous industries. In this work, we present a two-stage method for creating PFDs and PIDs, particularly for the large-scale synthesis of novel chemicals. It involves a sequential and complementary knowledge acquisition phase, followed by a knowledge processing and generation phase. The knowledge acquisition phase populates the knowledge base (providing a foundation) used in the subsequent generation phase. Specifically: (a) Agentic web navigation \cite{putta2024agentqadvancedreasoning, he2024webvoyager, abuelsaad2024-agente}, utilizing search engines and targeted queries, serves as a valuable, albeit incomplete, resource for accessing publicly available knowledge on PFDs and PIDs for well-known chemical processes. However, proprietary information regarding optimized process designs and highly specific control schemes is rarely publicly available, and the available information often represents simplified versions of real-world processes. (b) The knowledge processing and generation phase employs Retrieval-Augmented Generation (RAG) \cite{lewis2020retrieval}, leveraging pretrained large language models (LLMs) as computational engines to create high-quality, regulatory-compliant PFDs and PIDs for the large-scale synthesis of scarce or nonexistent chemicals. It leverages both external structured knowledge (from the acquisition phase) and the in-context learning capabilities of LLMs to automate PFD and PID generation. Furthermore, we address traditional open-domain question-answering (ODQA) tasks by providing precise textual answers to natural language questions (including factual, procedural, interpretive, and inferential questions) about PFD and PID generation. The multi-agent framework for autonomous web navigation centers around a meta-agent that coordinates specialized sub-agents (image, scholar, patent, wiki, and web insights agents) to gather comprehensive information about chemical process PFDs and PIDs. Each sub-agent uses SerpAPI to perform web searches within its specific domain and leverages language models to process and synthesize the retrieved information. The framework also includes a feedback loop that incorporates human experts and AI judges for quality control through self-reflection. (2) A Graph RAG framework organizes the generated knowledge into an ontological knowledge graph for improved information retrieval and question answering, enabling the automatic generation of PFDs and PIDs.  Figures \ref{fig:figure4} and \ref{fig:figure5} illustrate the two-step framework: autonomous agentic web navigation for high-fidelity data retrieval and synthesis, and a Graph RAG approach structuring this knowledge via ontological knowledge graphs to automatically generate PFDs and PIDs and facilitate complex ODQA tasks.

\vspace{-3mm}

\vspace{-2mm}
\section{Agentic Web Navigation Framework} %
We present a multi-agent framework for autonomous web navigation, optimized for knowledge creation. This framework enables independent web browsing, relevant information gathering, and insight synthesis related to PFDs and PIDs in chemical processes using online resources. It includes a meta-agent (or top-level agent) that orchestrates various specialized sub-agents. The meta-agent initiates the process by autonomously formulating queries to generate comprehensive knowledge about each chemical's PFDs and PIDs. By utilizing specialized sub-agents, each an expert in specific web-based information retrieval tasks, the meta-agent gathers and integrates detailed information from multiple sources. Each sub-agent uses SerpAPI to perform web searches and retrieve relevant information (e.g., images, scholarly articles, patents, Wikipedia, and web insights), which is then processed and synthesized by language models acting as computational engines to generate contextually relevant responses. The image sub-agent uses SerpAPI to extract and parse Google search results for relevant images of diagrams or flowcharts from publicly available sources. It then leverages LLMs to analyze the images and generate clear, concise, and contextually relevant responses. The scholar sub-agent uses SerpAPI to retrieve peer-reviewed scientific articles from Google Scholar. It then leverages LLMs to analyze, summarize, and synthesize the content, providing insightful responses to user queries. The patent sub-agent uses SerpAPI to retrieve patents and, with the help of LLMs, analyzes key sections to generate a concise overview, offering accurate and clear responses. The wiki sub-agent uses SerpAPI to retrieve relevant Wikipedia articles and then leverages LLMs to synthesize the information into a clear, concise overview, highlighting key facts and context to answer specific queries. The web insights sub-agent uses SerpAPI to gather insights from blogs, forums, and other relevant websites. It then leverages LLMs to analyze the content and provide accurate, concise responses by synthesizing data from retrieved sources and contextual knowledge. In short, to minimize redundant retrieval, each sub-agent is configured to handle unique data sources or content types, dynamically adapting its retrieval queries to ensure efficiency. The meta-agent decomposes the overall task \( Q \) into subtasks \( \{q_1, q_2, \dots, q_n\} \), such as retrieving images, searching for scholarly articles, exploring patents, gathering Wikipedia knowledge, and collecting general web data. This approach enables the meta-agent to prioritize subtasks by relevance, allowing sub-agents to work in parallel and speeding up retrieval. For each subtask \( q_i \), the meta-agent selects a sub-agent \( t_j \) by maximizing the cosine similarity between the text embeddings of the subtask query \( q_i \) and the documented capabilities \( d_j \) of each sub-agent. This similarity is defined as:

\vspace{-1mm}
\resizebox{0.95\linewidth}{!}{
\begin{minipage}{\linewidth}
\begin{equation}
\text{sim}(v(q_i), v(d_j)) = \frac{v(q_i) \cdot v(d_j)}{\|v(q_i)\| \|v(d_j)\|} \nonumber
\end{equation}
\end{minipage}
}

where \( v(q_i) \) and \( v(d_j) \) are the vector representations of the subtask query and the documented capabilities. The selection process is formalized as:

\resizebox{0.985\linewidth}{!}{
\begin{minipage}{\linewidth}
\begin{equation}
\text{SelectAgent}(q_i) = \arg\max_{t_j} \text{sim}(v(q_i), v(d_j)) \nonumber
\end{equation}
\end{minipage}
}

where \( t_j \) is the sub-agent, and \( d_j \) describes the agent’s relevance to the task. The documented capabilities of sub-agents \( d_j \) include specific skills, such as retrieving images or conducting literature searches. This documentation outlines their data sources—platforms like Google Images, Google Scholar, and patent databases (e.g., USPTO and EPO)—enabling the meta-agent to select the most relevant sub-agent for each subtask and optimize information retrieval. Dependencies between subtasks are managed through a Directed Acyclic Graph (DAG) \( \mathcal{G} = (\mathcal{V}, \mathcal{E}) \), where nodes \( \mathcal{V} \) represent subtasks \( q_i \), and edges \( \mathcal{E} \) indicate dependencies \( e_{ij} \) between them. This structure ensures the correct order of subtasks: if subtask \( q_i \) depends on \( q_j \), then \( q_j \) must be completed before \( q_i \) can begin. Independent subtasks can be executed simultaneously, enhancing overall efficiency. After the meta-agent identifies the most suitable expert sub-agent for each subtask, it invokes the sub-agent with parameters: \( p_1 \) (Instructions), \( p_2 \) (Context), and \( p_3 \) (Specific Query). This invocation is represented as:

\vspace{-1mm}
\resizebox{0.985\linewidth}{!}{
\begin{minipage}{\linewidth}
\begin{equation}
t_j(p_1, p_2, p_3)  \nonumber
\end{equation}
\end{minipage}
}

Each sub-agent synthesizes information, identifies patterns, and extracts key insights from multiple sources. For the image sub-agent, a set of images \( I = \{i_1, i_2, \dots, i_k\} \) is retrieved, and their embeddings \( v(I) = \{v(i_1), v(i_2), \dots, v(i_k)\} \) are computed using the CLIP model~\cite{radford2021learning}. The cosine similarity between each image embedding and the query embedding \( v(q_i) \) is calculated as follows:

\vspace{-1mm} 
\resizebox{0.95\linewidth}{!}{ 
\begin{minipage}{\linewidth} 
\begin{equation} 
\text{sim}(v(I), v(q_i)) = \frac{v(I) \cdot v(q_i)}{|v(I)| |v(q_i)|} \nonumber
\end{equation} 
\end{minipage} 
}

A selected subset \( I_J \subset I \) is presented to the multimodal LLM, which interprets the visual information and generates a coherent summary \( D_I \) relevant to the context of PFDs and PIDs based on the query \( q_i \). Similarly, the scholar sub-agent processes scientific articles \( A = \{a_1, a_2, \dots, a_m\} \), embedding them as \( v(A) = \{v(a_1), v(a_2), \dots, v(a_m)\} \), and calculates the cosine similarity with the query embedding \( v(q_{i}) \) to identify relevant articles. A selected subset is then input into the LLM to generate output \( D_A \). For Wikipedia knowledge retrieval, specific web pages \( W = \{w_1, w_2, \dots, w_p\} \) are embedded as \( v(W) = \{v(w_1), v(w_2), \dots, v(w_p)\} \), and relevant articles are identified through cosine similarity and input into the LLM to produce output \( D_W \). For patent retrieval, specific patents \( P = \{p_1, p_2, \dots, p_r\} \) are embedded as \( v(P) = \{v(p_1), v(p_2), \dots, v(p_r)\} \), and relevant patents are identified using cosine similarity with the query embedding \( v(q_i) \). A selected subset is input into the LLM to generate a coherent output \( D_P \). Lastly, the general web data gathering sub-agent processes insights \( G = \{g_1, g_2, \dots, g_q\} \), such as blog posts and reports on PFDs and PIDs, embedding them as \( v(G) = \{v(g_1), v(g_2), \dots, v(g_q)\} \). Relevant insights are selected for the LLM to generate a summarized output \( D_G \). The meta-agent aggregates these outputs into a coherent response \( A \):

\vspace{-1mm} 
\resizebox{0.985\linewidth}{!}{ 
\begin{minipage}{\linewidth} 
\begin{equation}
A = \text{MetaAgent}_{\text{LLM}}(D_I, D_A, D_P, D_W, D_G) \nonumber
\end{equation} 
\end{minipage} 
}

This structured workflow enhances the efficiency and effectiveness of information retrieval and knowledge acquisition by detailing task delegation, execution, and aggregation. To ensure the accuracy and relevance of the generated content, the framework implements an iterative feedback loop in which human experts, a gold-standard language model (Gold-LLM-as-a-Judge), and a reward model (Reward-Model-as-Judge) provide feedback \( F_i \). This reflective feedback process enables the meta-agent to evaluate its performance and refine its outputs, fostering continuous improvement. The iterative process continues until the output meets specified quality standards or until a maximum number of iterations \( N_{\text{max}} \) is reached, represented as:

\vspace{-1mm} 
\resizebox{0.985\linewidth}{!}{ 
\begin{minipage}{\linewidth} 
\begin{equation}
A_{i+1} = \text{MetaAgent}_{\text{LLM}}(Q, A_i, F_i) \nonumber
\end{equation} 
\end{minipage} 
}

\vspace{1mm}
The goal is to produce an optimal output \( A^* \) that maximizes \( P(A \mid Q, T, D, \theta) \). Here, \(\theta\) represents the meta-agent's associated language model parameters, which encapsulate its pre-trained knowledge and enable effective information processing. \( T = \{t_1, t_2, \dots, t_n\} \) is the pool of specialized sub-agents available for task execution, and \( D = \{d_1, d_2, \dots, d_n\} \) represents the set of documented capabilities of all sub-agents. The autonomous knowledge generation process is described in \textbf{Algorithm~\ref{alg:autonomous_navigation_framework}}, which outlines the framework for agent-based web navigation. An ontological knowledge graph is constructed based on the generated knowledge to organize and retrieve relevant information within the Graph RAG framework. This construction facilitates the generation of more accurate, context-aware, and efficient responses. These capabilities are particularly valuable in complex and specialized domains, such as the generation of PFDs and PIDs for chemical processes ODQA tasks.

\vspace{-1mm} 
\section{Graph Retrieval-Augmented Generation}
Graph RAG surpasses traditional RAG by leveraging knowledge graphs to overcome the limitations of traditional RAG in handling complex queries. While traditional RAG excels at retrieving simple, isolated facts, it often struggles to synthesize information across multiple sources for complex multi-hop reasoning tasks. Graph RAG, on the other hand, utilizes the structured relationships within knowledge graphs to traverse and reason effectively, providing deeper insights and more comprehensive responses. The process begins by constructing unimodal (text-only) knowledge graphs (KGs) through parsing unstructured documents, extracting relevant information, and structuring it into triples (subject-predicate-object). To facilitate this extraction and structuring, the documents are divided into manageable chunks using a sliding window technique. Let each document in a set of \( M \) documents be denoted as \( D^{(m)} \), where \( m = 1, 2, \dots, M \) represents the document index. Each document \( D^{(m)} \) is divided into chunks based on a token window size. Each chunk is denoted as \( C_i^{(m)} \), where \( i = 1, 2, \dots, N^{(m)} \) denotes the chunk index within document \( D^{(m)} \), and \( N^{(m)} \) is the number of chunks in that document, as defined below:

\resizebox{0.985\linewidth}{!}{ 
\begin{minipage}{\linewidth} 
\begin{equation}
C_i^{(m)} = D^{(m)}\left[(i - 1) \cdot s : (i - 1) \cdot s + w\right] \nonumber
\end{equation}
\end{minipage} 
}

Here, \( C_i^{(m)} \) represents the \( i \)-th chunk in document \( D^{(m)} \). The window size \( w \) (in tokens) and the stride \( s \) (the step size between consecutive chunks, also in tokens) are applied consistently across all documents. Choosing an optimal window size \( w \) and stride \( s \) requires balancing granularity with context retention. A smaller window with a smaller stride (resulting in higher overlap) may capture more detailed information but increase the computational cost of processing by language models. In contrast, a larger window size with a larger stride may better preserve context but dilute specific details. Although the sliding window technique helps maintain context, critical information near chunk boundaries may still be fragmented, impacting coherence and completeness, which can reduce retrieval effectiveness. To address the limitations, we propose a hybrid approach that combines the sliding window technique with a content-aware method \cite{anthropic_contextual_retrieval}. In this approach, a language model generates a contextual description for each chunk \( C_i^{(m)} \), denoted \( \text{ctx}_i^{(m)} \), which provides a concise summary linking the chunk to the document’s broader content, thereby enhancing retrieval relevance. We then prepend this context to the original chunk before encoding:

\resizebox{0.985\linewidth}{!}{ 
\begin{minipage}{\linewidth} 
\begin{equation}
C_i^{\prime (m)} = \text{ctx}_i^{(m)} \oplus C_i^{(m)} \nonumber
\end{equation}
\end{minipage} 
}

where \( \oplus \) denotes the concatenation operator. This approach embeds additional context, retains essential information, and improves retrieval accuracy. After enriching the chunks with context, they are used to construct the knowledge graph. To construct a knowledge graph \( \mathcal{G} = (\mathcal{V}, \mathcal{E}) \), it is essential to extract and represent entities and their relationships in a structured format. The set of entities \( \mathcal{V} \) serves as the nodes, while the relationships between entities, represented as \( \mathcal{E} \), form the directed edges. This structured representation enables the encoding of semantic information, providing a robust framework for advanced reasoning and retrieval tasks. We extract both entities and relationships using language models. The process begins with Named Entity Recognition (NER) to identify and extract entities within each chunk, resulting in a set of entities denoted as:

\resizebox{0.985\linewidth}{!}{ 
\begin{minipage}{\linewidth} 
\begin{equation}
E_i^{(m)} = \{ e_j^{(i, m)} : j = 1, 2, \dots, K^{(m, i)} \} \nonumber
\end{equation}
\end{minipage} 
}

where \( e_j^{(i, m)} \) represents the \( j \)-th entity extracted from the \( i \)-th chunk \( C_i^{(m)} \) of document \( D^{(m)} \), and \( K^{(i, m)} \) is the total number of entities in this chunk. Next, Relation Extraction (RE) identifies relationships between pairs of entities within the same chunk. These relationships are represented as:

\resizebox{0.985\linewidth}{!}{ 
\begin{minipage}{\linewidth} 
\begin{equation}
R_i^{(m)} = \{ (e_j^{(i, m)}, r^{(i, m)}_{j,k}, e_k^{(i, m)}) : e_j^{(i, m)}, e_k^{(i, m)} \in E_i^{(m)} \} \nonumber
\end{equation}
\end{minipage} 
}

Here, each triple \( (e_j^{(i, m)}, r^{(m, i)}_{j,k}, e_k^{(i, m)}) \) denotes a relation \( r^{(i, m)}_{j,k} \) between the entities \( e_j^{(i, m)} \) and \( e_k^{(i, m)} \).
To improve search accuracy, we merge duplicate entities by identifying and combining those that represent the same concepts but appear differently in the data. Each entity \( e_j^{(i, m)} \) is represented by a vector embedding \( \mathbf{v}_{j}^{(i, m)} \), generated using OpenAI's text-embedding-3-small model, which captures the entity's semantic meaning. The semantic similarity between two entities \( e_j^{(i, m)} \) and \( e_{j'}^{(i', m')} \) is calculated using cosine similarity:

\vspace{-3mm}
\resizebox{0.875\linewidth}{!}{ 
\begin{minipage}{\linewidth} 
\begin{equation}
\text{sim}(\mathbf{v}_j^{(i, m)}, \mathbf{v}_{j'}^{(i', m')}) = \frac{\mathbf{v}_j^{(i, m)} \cdot \mathbf{v}_{j'}^{(i', m')}}{\|\mathbf{v}_j^{(i, m)}\| \|\mathbf{v}_{j'}^{(i', m')}\|} \nonumber
\end{equation}
\end{minipage} 
}

If the similarity exceeds a threshold \( \tau_{\text{sim}} \), the entities are considered semantically similar. For further refinement, we use the Levenshtein distance to compute a string similarity ratio:

\vspace{-3mm}
\resizebox{0.875\linewidth}{!}{ 
\begin{minipage}{\linewidth} 
\begin{equation}
\text{str\_sim}(e_j^{(i, m)}, e_{j'}^{(i', m')}) = 1 - \frac{d_{\text{lev}}(e_j^{(i, m)}, e_{j'}^{(i', m')})}{\max(|e_j^{(i, m)}|, |e_{j'}^{(i', m')}|)} \nonumber
\end{equation}
\end{minipage} 
}

where \( d_{\text{lev}} \) denotes the Levenshtein distance between the two entity strings. An entity pair is deemed a duplicate and merged if both the semantic similarity \( \text{sim}(\mathbf{v}_{j}^{(i, m)}, \mathbf{v}_{j'}^{(i', m')}) \geq \tau_{\text{sim}} \) and the string similarity \( \text{str\_sim}(e_j^{(i, m)}, e_{j'}^{(i', m')}) \geq \tau_{\text{str}} \) satisfy predefined thresholds. We apply the hierarchical Leiden algorithm to detect communities \( \mathcal{C}_i \) at multiple levels of granularity, maximizing the modularity \( Q_{\text{Mod}} \) of the knowledge graph. Modularity \( Q_{\text{Mod}} \) quantifies the quality of the community structure by measuring the density of connections within communities compared to those between communities, with higher values indicating denser internal connections and sparser external connections. It is defined as:

\vspace{0mm}
\resizebox{0.95\linewidth}{!}{ 
\begin{minipage}{\linewidth} 
\begin{equation}
Q_{\text{Mod}} = \frac{1}{2m} \sum_{i,j} \left[ A_{ij} - \frac{k_i k_j}{2m} \right] \delta(c_i, c_j) \nonumber
\end{equation}
\end{minipage} 
}

Here, \( A_{ij} \) represents the adjacency matrix, indicating the presence of an edge between nodes \( i \) and \( j \); \( k_i \) and \( k_j \) are the degrees of nodes \( i \) and \( j \); \( m \) is the total number of edges in the graph, used to normalize the adjacency matrix to reflect the expected density of connections in a random graph (where node degrees are preserved, but specific connections are randomized); and \( \delta(c_i, c_j) \) is the Kronecker delta function, equal to 1 if nodes \( i \) and \( j \) belong to the same community and 0 otherwise. The hierarchical Leiden algorithm partitions the graph into \( L \) communities, \( \{\mathcal{C}_1, \dots, \mathcal{C}_L\} \), by maximizing \( Q_{\text{Mod}} \), revealing a strong community structure. The process involves three steps: (1) the local moving phase, where nodes are shifted between communities to maximize modularity; (2) the aggregation phase, where detected communities are merged into super-nodes; and (3) repetition of these phases until no further improvement in modularity is possible. Each community \( \mathcal{C}_i = (\mathcal{V}_{\mathcal{C}_i}, \mathcal{E}_{\mathcal{C}_i}) \) consists of its nodes \( \mathcal{V}_{\mathcal{C}_i} \) and edges \( \mathcal{E}_{\mathcal{C}_i} \). The algorithm facilitates focused retrieval in knowledge graphs by detecting communities and refining them to align with query-specific subgraphs. For complex multi-hop reasoning tasks, relevant information often spans multiple communities. To address this, our approach prioritizes the top-\( K \) communities \( \{\mathcal{C}_1, \mathcal{C}_2, \dots, \mathcal{C}_K\} \), ranked based on the cosine similarity between the user query \( Q \) and summaries of relationship paths within each community. Each community \( \mathcal{C}_i \) is summarized using a language model that condenses relationship paths \( R_i \):

\vspace{-1mm}
\resizebox{0.95\linewidth}{!}{ 
\begin{minipage}{\linewidth} 
\begin{equation}
S_i = \text{LLM}(R_i) = \arg\max_{S} \, P(S \mid R_i) \nonumber
\end{equation}
\end{minipage} 
}

where \( S_i \) is the summary, and \( P(S \mid R_i) \) represents the likelihood of \( S \) given \( R_i \), encompassing both direct and multi-hop relationships. These summaries are converted into vector embeddings \( \mathbf{v}(S_i) \) using a text-embedding model, enabling efficient similarity computation with the embedded query \( \mathbf{v}(Q) \):

\vspace{-2mm}
\resizebox{0.935\linewidth}{!}{ 
\begin{minipage}{\linewidth} 
\begin{equation}
d(Q, \mathcal{C}_i) = \frac{\langle \mathbf{v}(Q), \mathbf{v}(S_i) \rangle}{\|\mathbf{v}(Q)\| \cdot \|\mathbf{v}(S_i)\|} \nonumber
\end{equation}
\end{minipage} 
}

The top-\( K \) communities with the highest similarity scores are selected, ensuring that a diverse and relevant set of subgraphs is aggregated. These selected communities are then combined into a query-specific subgraph \( G_Q = (\mathcal{V}_Q, \mathcal{E}_Q) \), defined as:

\vspace{-3mm}
\resizebox{0.935\linewidth}{!}{ 
\begin{minipage}{\linewidth} 
\begin{equation}
\mathcal{V}_Q = \bigcup_{i=1}^{K} \mathcal{V}_{\mathcal{C}_i}, \quad \mathcal{E}_Q = \bigcup_{i=1}^{K} \mathcal{E}_{\mathcal{C}_i} \nonumber
\end{equation}
\end{minipage} 
}

This subgraph retains critical nodes \( \mathcal{V}_Q \) and edges \( \mathcal{E}_Q \) from the top-\( K \) communities, capturing the essential connections and semantic relationships needed to address the query. Paths \( P = (e_1, r_1, e_2, \dots, e_k) \), where \( e_i \) are entities and \( r_i \) are relationships, are extracted from \( G_Q \). These paths encode semantic dependencies vital for reasoning. A language model generates the answer \( A \) by integrating the query \( Q \) and paths \( P \):

\resizebox{0.95\linewidth}{!}{ 
\begin{minipage}{\linewidth} 
\begin{equation}
A = \text{LLM}(Q, P) = \arg\max_{A} \, P(A \mid Q, P) \nonumber
\end{equation}
\end{minipage} 
}

where \( P(A \mid Q, P) \) denotes the probability of \( A \) given \( Q \) and \( P \), ensuring that the response is contextually accurate and aligned with the query's intent. The construction and utilization of a knowledge graph for the ODQA task are detailed in \textbf{Algorithm~\ref{alg:knowledge_graph_ODQA}}.

\section{Experiments}

\vspace{-1mm}
\subsection{Datasets}
We constructed a comprehensive dataset of over 1,070 chemicals with significant applications across diverse industries, including electronics manufacturing, oil and gas, pharmaceuticals, renewable energy, chemical production, mining, water treatment, and food and beverage. This dataset, meticulously curated from the product catalogs of leading chemical manufacturers like BASF, Dow Chemicals, and DuPont, ensures reliability and consistency by grounding the data in credible sources, thereby minimizing ambiguities and inaccuracies associated with unstructured, free-form inputs. It comprises two subsets: a primary subset of 1,020 chemicals used for autonomous web navigation, domain-specific data retrieval, and generating PFDs and PIDs to construct ontological knowledge graphs as foundational databases; and a secondary evaluation subset of 50 chemicals to rigorously assess the framework's robustness and generalizability in auto-generating PFDs and PIDs. Additionally, we developed a custom ODQA dataset of 6,000 QA pairs focused on process diagrams of the primary subset, with questions systematically generated using predefined templates and answers produced by benchmark LLMs like GPT-4o. Covering categories such as fact-based, logical, comparative, causal, operational, multi-hop, and procedural questions, each QA pair underwent meticulous validation before inclusion, enabling the evaluation of the framework's accuracy, contextual relevance, and capability to address diverse technical queries.

\vspace{-1mm}
\subsection{Experimental Settings}
In our work, we utilized SmolLM2-360M-Instruct [Allal et al., 2024], a pre-trained and fine-tuned model, as the computational engine to compare the performance of Graph RAG with a pre-trained LLM (leveraging ontological knowledge graphs constructed from PFD and PID descriptions of the primary subset of chemicals) and a fine-tuned LLM without Graph RAG (instruct-tuned using the same knowledge base). This comparison aimed to evaluate the impact of Graph RAG on pre-trained model performance. Graph RAG with a pre-trained LLM incurs higher computational costs due to graph construction and retrieval, resulting in slower overall inference times because of graph traversal. However, the pre-trained LLM processes retrieved context efficiently. It excels in dynamic knowledge updates and multi-hop reasoning. In contrast, a fine-tuned LLM without Graph RAG offers faster inference by embedding retrieval within the model but requires costly fine-tuning and is less flexible for adapting to new data without re-fine-tuning. The fine-tuned LLM’s inference time is slightly higher due to its larger, task-specific parameterization. We also utilized OpenAI's text-embedding-3-small model for text encoding tasks and CLIP embeddings for multimodal processing, enabling the encoding of visual data. The framework integrated the Graph RAG approach with Neo4j to facilitate structured knowledge retrieval. To fine-tune SmolLM2-360M-Instruct, we generated an instruction-following dataset for the task of producing descriptions of PFDs and PIDs for the primary subset of chemicals. The dataset was structured with instruction prompts and corresponding target outputs and was used for supervised fine-tuning to optimize the model's task-specific performance. Note: For Graph RAG with pre-trained LLMs, the same knowledge—PFD and PID descriptions of the primary subset of chemicals—was used for constructing the knowledge graph and did not involve fine-tuning the LLM itself. The fine-tuning process employed advanced techniques such as gradient accumulation and Low-Rank Adaptation (LoRA) for efficient training. Key hyperparameters included a learning rate of \(5 \times 10^{-5}\), a batch size of 8 per device, 3 training epochs, and 4 gradient accumulation steps, simulating an effective batch size of 32 (8 × 4). This configuration allowed the model to perform updates as if it had processed 32 samples in one step, even though only 8 samples were loaded into memory at a time. LoRA reduced trainable parameters by updating low-rank matrices (e.g., rank \(r = 4\), scaling factor \(\alpha = 16\)) and applied dropout (e.g., 0.1) to updates for regularization. LoRA fine-tuned only specific layers, such as attention weights, while typically freezing biases, and integrated seamlessly with gradient accumulation. Low-rank matrices were initialized randomly to ensure efficient convergence, while regularization prevented overfitting. We employed LoRA with mixed precision (FP16), further optimizing memory usage and training speed. Rigorous evaluation strategies, such as per-epoch validation using metrics like loss (e.g., cross-entropy), ensured scalable and effective fine-tuning tailored to specific tasks. All experiments were conducted using NVIDIA Tesla T4 GPUs for efficient computation, and the framework was implemented in Python with PyTorch and Unsloth.

\vspace{-1mm}
\subsection{Experimental Studies}
The framework's performance was evaluated across multiple tasks: (a) comparing auto-generated PFDs and PIDs for chemicals in the secondary evaluation subset to ground-truth data; (b) evaluating its ability to answer diverse queries from the ODQA dataset, including logical, causal, procedural, and multi-hop reasoning questions; (c) analyzing the impact of structured retrieval on multi-hop reasoning and contextual accuracy by comparing Graph RAG combined with a pre-trained LLM to a fine-tuned LLM without Graph RAG, used as a baseline; and (d) demonstrating Graph RAG's superiority over traditional RAG in handling complex queries and generating accurate, context-aware responses.

\vspace{-1mm}
\subsection{Results}
In this section, we present the experimental results on knowledge generation. Additional experimental results are discussed in the appendix. Figure \ref{fig:figure6} illustrates the evaluation metrics for the autonomous agentic web navigation framework, which is designed to automate and optimize the processes of gathering and synthesizing information for PFDs and PIDs for the primary subset of chemicals from publicly available online sources. The framework's outputs were evaluated using the NVIDIA Nemotron-4-340B-Reward model across metrics such as helpfulness, correctness, coherence, complexity, and verbosity. Each attribute was evaluated on a four-point Likert scale (continuous), with 0 indicating the lowest quality and 4 indicating the highest. The results highlight the quality of the generated knowledge, particularly its coherence and accuracy. We computed text embeddings of chemical PFDs and PIDs generated by the agentic web navigation framework using the OpenAI text-embedding-3-small model, followed by clustering to identify the optimal number of groups. These clusters were then visualized using t-SNE and PCA plots.

\vspace{-3mm}
\begin{figure}[!ht]
    \centering
    \resizebox{0.90\linewidth}{!}{
        \hspace*{0mm}\includegraphics[keepaspectratio,trim=0.0cm 0.025cm 0cm 0.025cm,clip]{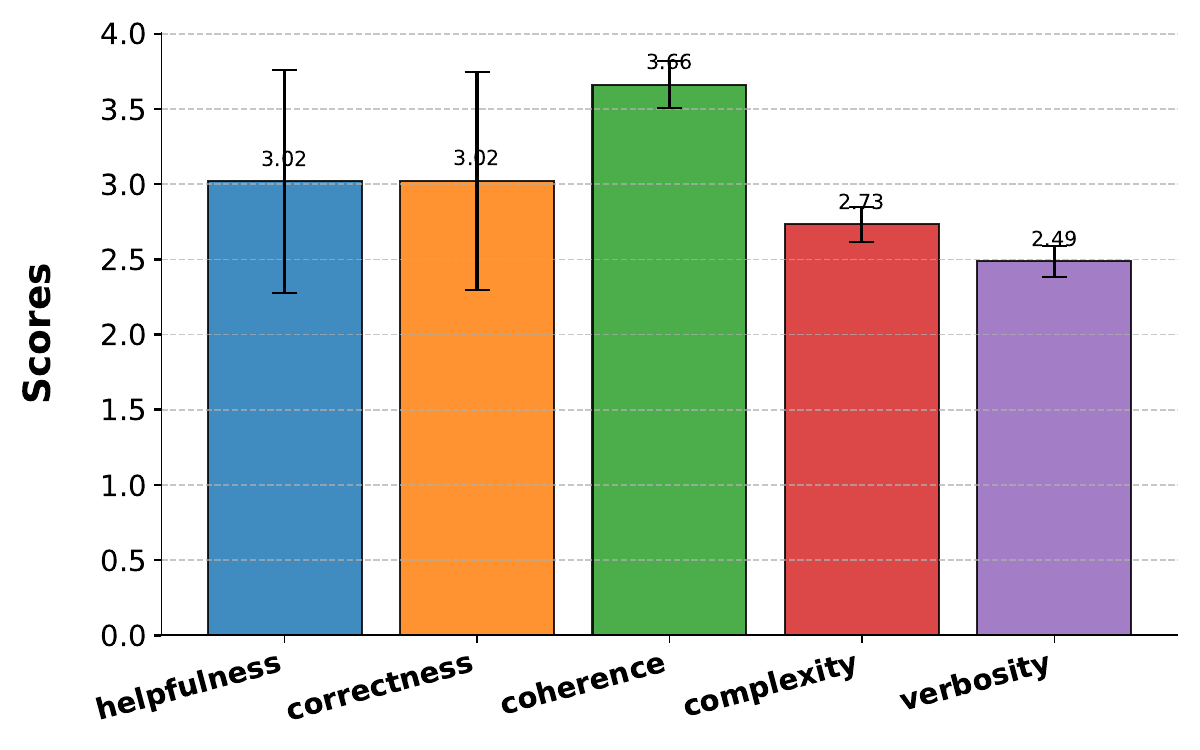}
    }
    \vspace{-2mm}
    \caption{The figure presents the evaluation results of the agentic web navigation framework in generating PFD and PID knowledge for chemical processes, benchmarked using NVIDIA Nemotron-4-340B-Reward model and scored on a scale from 0 to 4, where 0 represents the lowest and 4 the highest performance.}
    \label{fig:figure6}
    \vspace{-4mm}
\end{figure}

\vspace{-2mm}
\begin{figure}[!ht]
    \centering
    \begin{subfigure}[b]{0.8\linewidth}
        \centering
        \resizebox{\linewidth}{!}{
            \hspace*{0mm}\includegraphics[keepaspectratio,trim=0.0cm 0.025cm 0cm 0.025cm,clip]{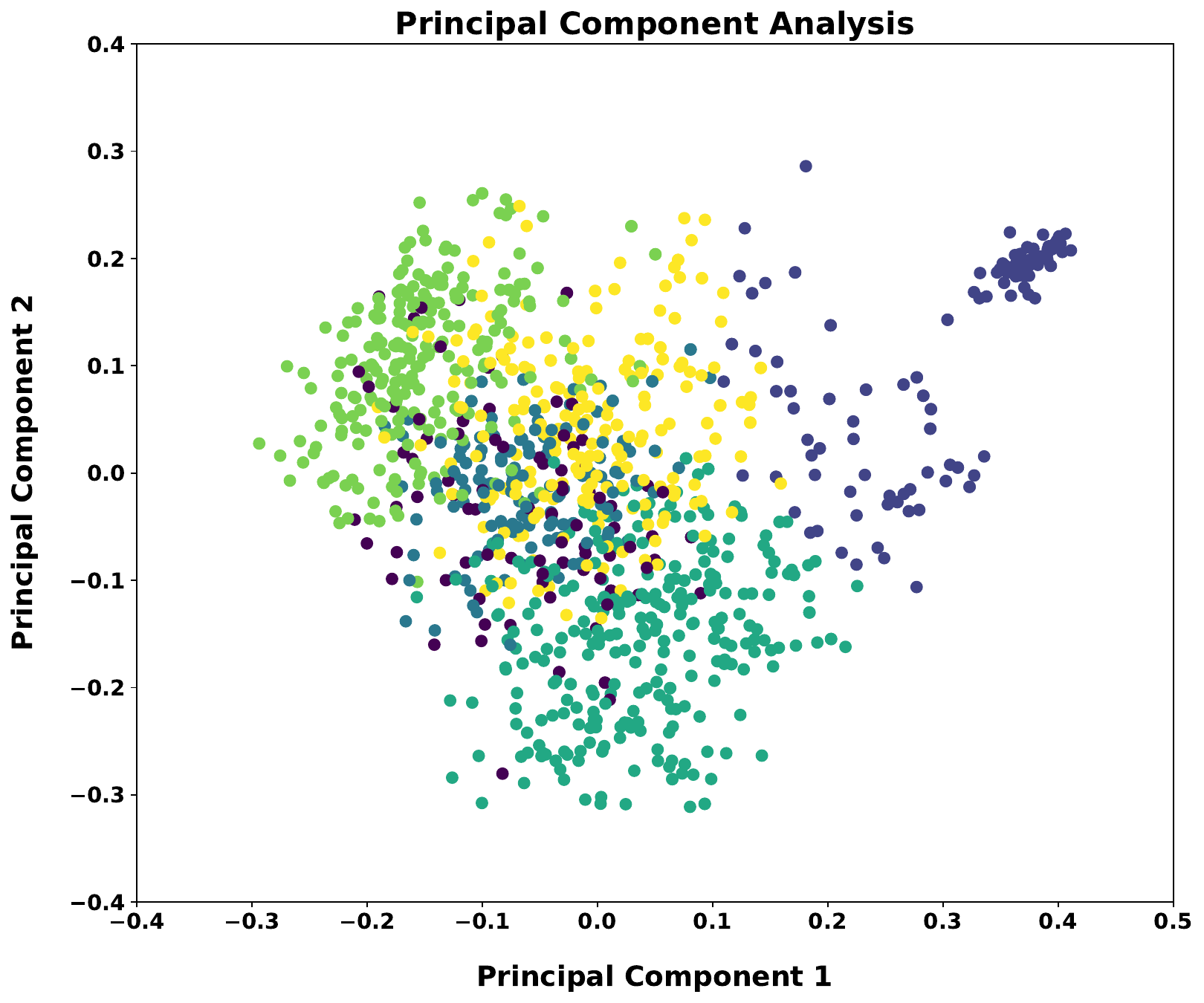}
        }
        \vspace{-4mm}
        \caption{PCA plot description.}
        \label{fig:figure7}
    \end{subfigure}
    
    \vspace{1mm}  
    
    \begin{subfigure}[b]{0.8\linewidth}
        \centering
        \resizebox{\linewidth}{!}{
            \hspace*{0mm}\includegraphics[keepaspectratio,trim=0.0cm 0.025cm 0cm 0.025cm,clip]{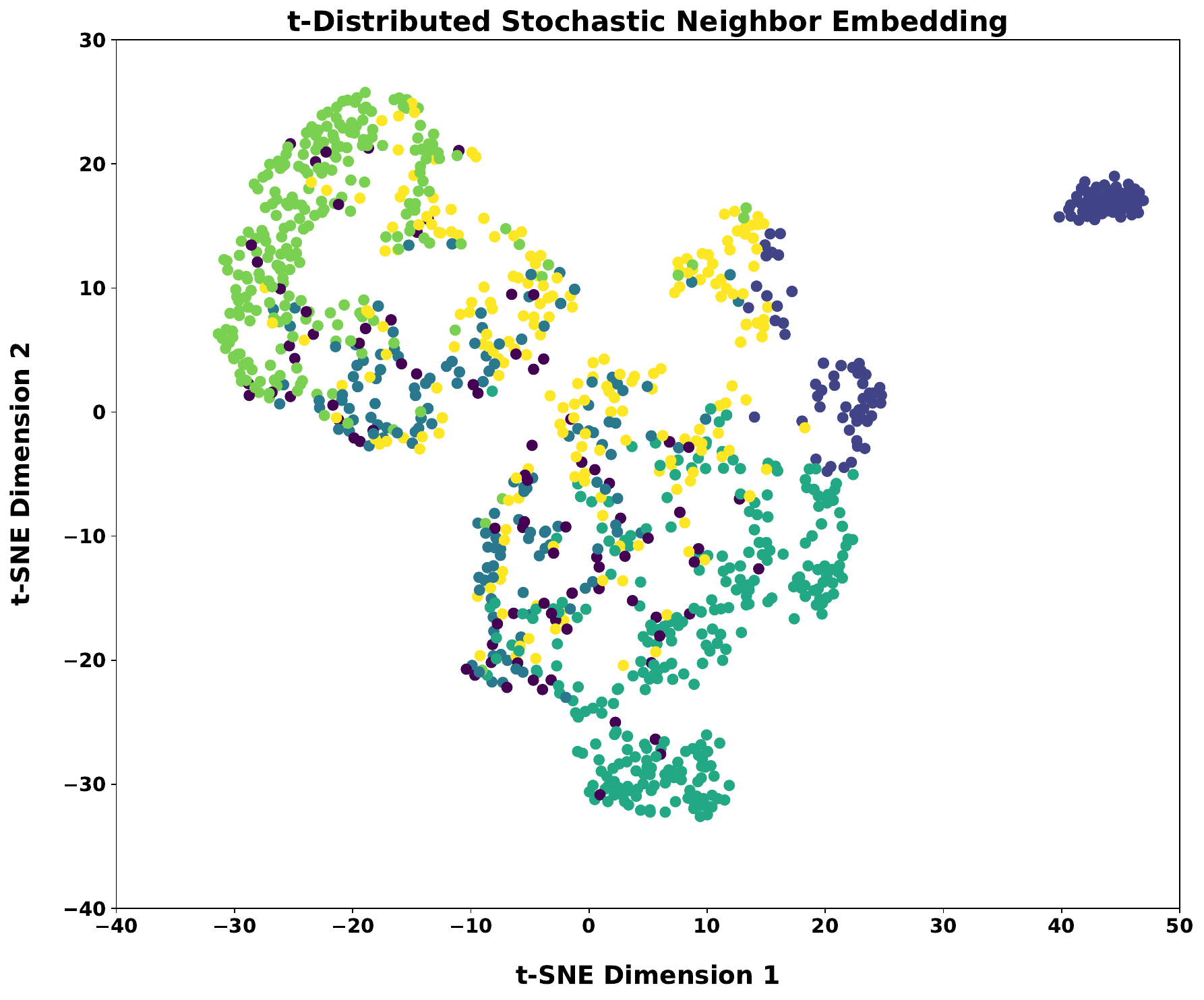}
        }
        \vspace{-4mm}
        \caption{t-SNE plot description.}
        \label{fig:figure8}
    \end{subfigure}    
    \vspace{-3mm}
    \caption{The figure shows the PCA projection and t-SNE visualization of text-level embeddings for PFDs and PIDs.}
    \vspace{-3mm}
\end{figure}

\begin{figure}[!ht]
    \centering
        \centering
        \resizebox{0.8\linewidth}{!}{
            \hspace*{0mm}\includegraphics[keepaspectratio,trim=0.0cm 0.025cm 0cm 0.025cm,clip]{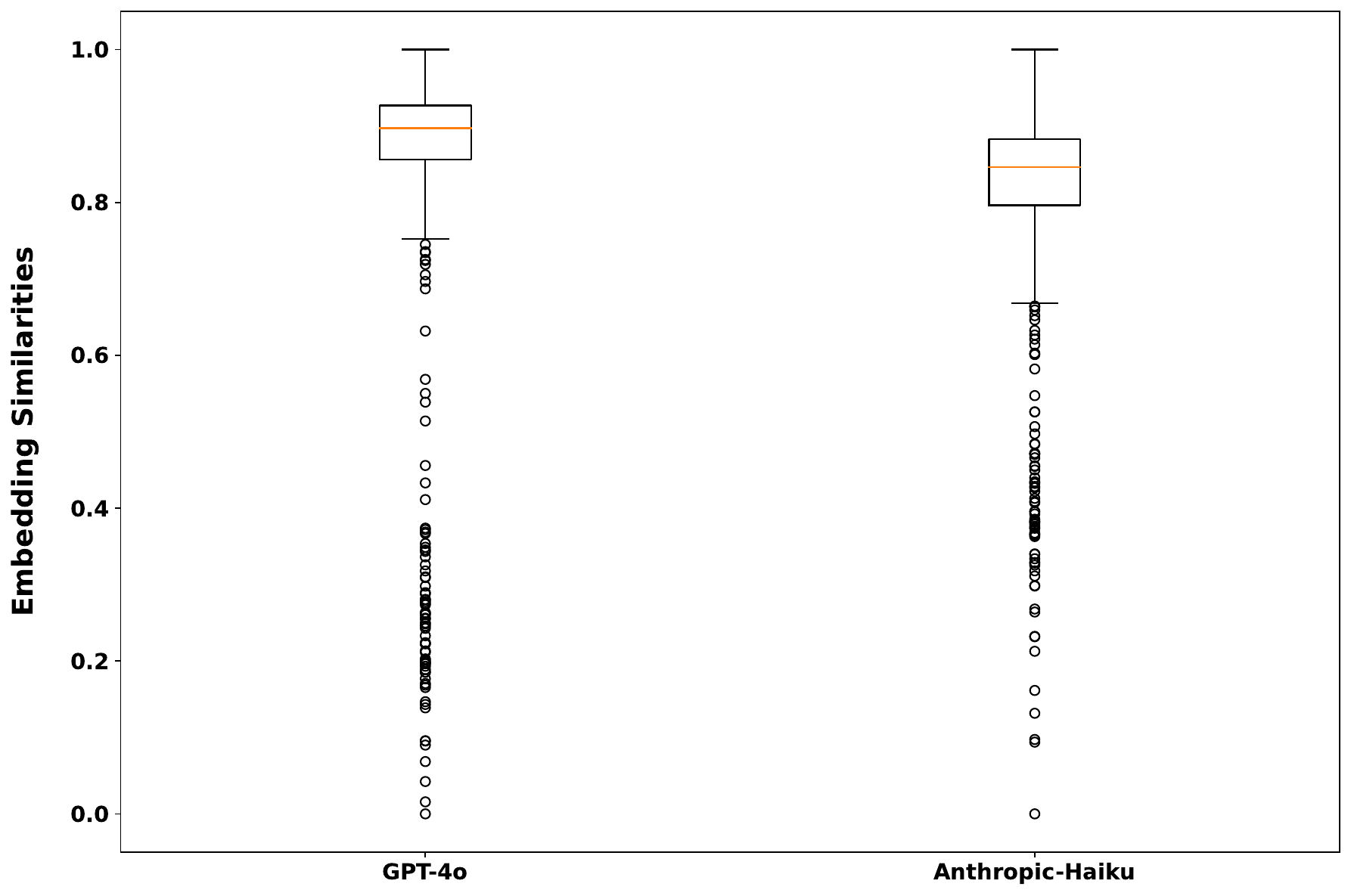}
        }
        \vspace{-2mm}
        \caption{The figure shows a box-and-whisker plot comparing the semantic closeness of knowledge generated by the web navigation framework, GPT-4o, and Anthropic-Haiku for PFDs and PIDs of chemical processes.}
    \label{fig:figure9}
    \vspace{-6mm}
\end{figure}

\begin{figure}[!ht]
    \centering
    \vspace{2mm}  
        \begin{subfigure}[b]{0.8\linewidth}
        \centering
        \resizebox{\linewidth}{!}{
            \hspace*{0mm}\includegraphics[keepaspectratio,trim=0.0cm 0.025cm 0cm 0.025cm,clip]{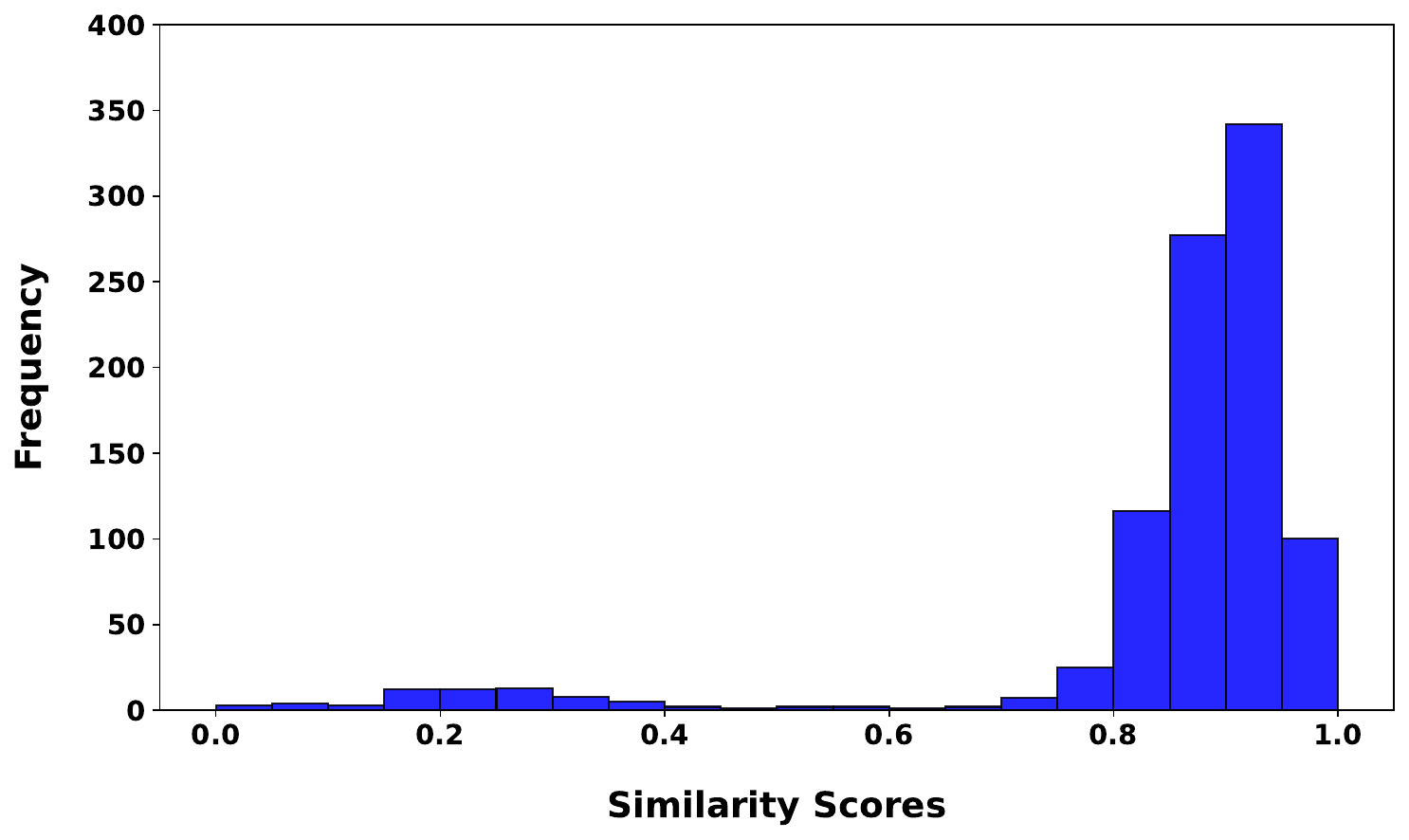}  
        }
        \vspace{-4mm}
        \caption{GPT-4o Vs web knowledge}
        \label{fig:figure10}
    \end{subfigure}
    \vspace{2mm}  
    \begin{subfigure}[c]{0.8\linewidth}
          \centering
            \resizebox{\linewidth}{!}{
            \hspace*{0mm}\includegraphics[keepaspectratio,trim=0.0cm 0.025cm 0cm 0.025cm,clip]{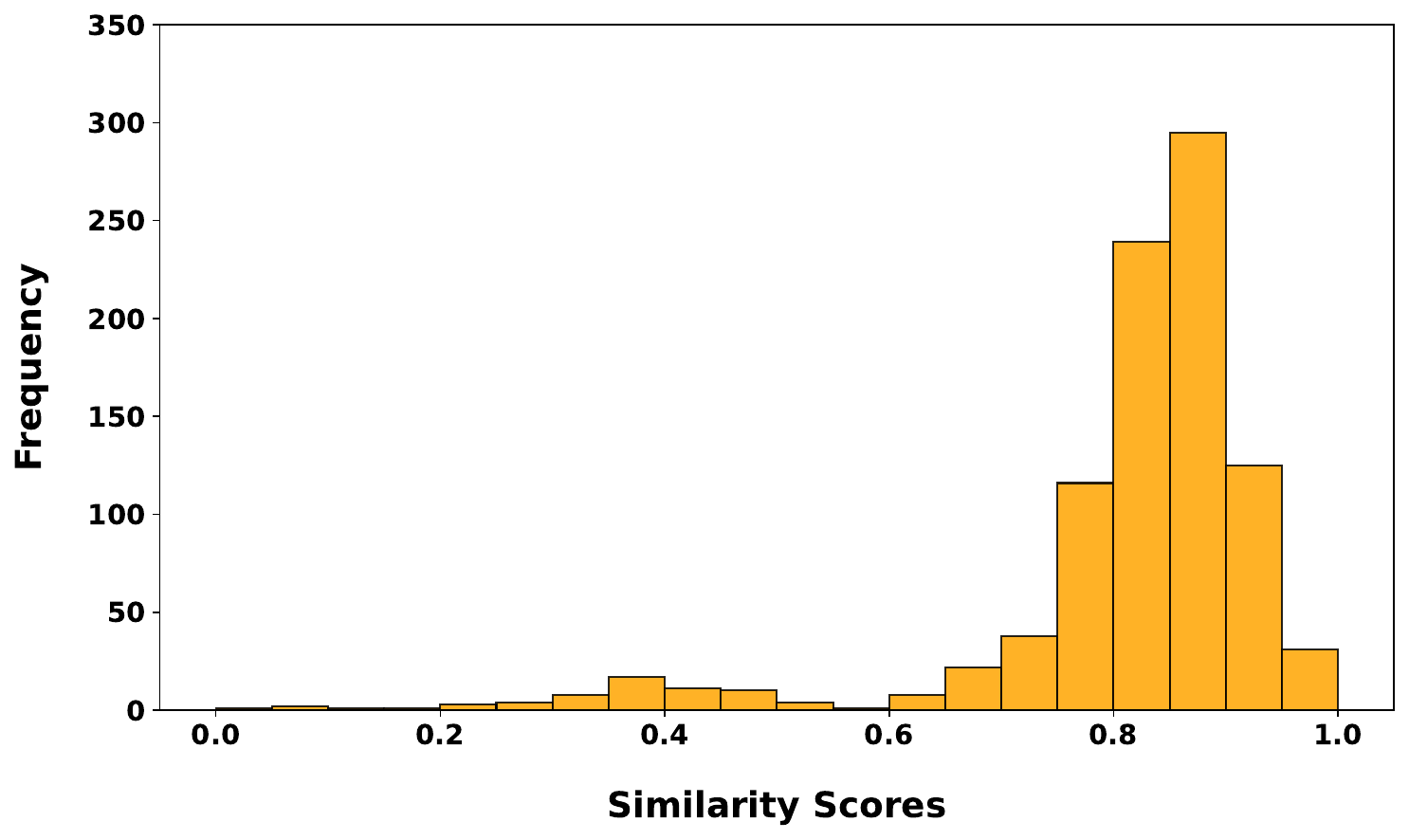}  
        }
        \vspace{-4mm}
        \caption{Anthropic-Haiku Vs web knowledge.}
        \label{fig:figure11}
    \end{subfigure}   
    \vspace{-4mm}
    \caption{The figure shows histograms of similarity scores comparing GPT-4o and Anthropic-Haiku-generated knowledge with web knowledge.}
    \vspace{-1mm}
\end{figure}

The t-SNE visualization (see Figure \ref{fig:figure7}) revealed distinct groupings, highlighting non-linear separations, while the PCA plot (see Figure \ref{fig:figure8}) displayed the variance structure with less pronounced clustering. PFDs and PIDs within a cluster share common characteristics, suggesting similar industrial designs, while those in different clusters exhibit distinct characteristics, reflecting diverse design strategies. This clustering analysis provides valuable insights for process optimization, design, and risk assessment. Figure \ref{fig:figure9} presents a box-and-whisker plot comparing the embedding similarities of knowledge generated by the agentic web navigation framework with two proprietary LLMs (GPT-4o and Anthropic-Haiku) for PFDs and PIDs of chemical processes. The plot suggests that the generated knowledge has a closer semantic relationship to text from GPT-4o compared to Anthropic-Haiku. However, outliers for both models indicate instances of weak similarity, suggesting differing strategies or biases in knowledge generation, particularly given the variability of available web information. Figures \ref{fig:figure10} and \ref{fig:figure11} display histograms comparing the similarity of knowledge generated by GPT-4o and Anthropic-Haiku to web knowledge. These plots reveal the frequency of various similarity scores, providing insights into the models' semantic alignment with web-sourced information. The distinct patterns in these distributions highlight variations in the models' alignment with web knowledge, with GPT-4o appearing more closely aligned compared to Anthropic-Haiku.

\vspace{-3mm}

\section{Conclusion}
In conclusion, this paper introduces an autonomous agentic web navigation framework that integrates GraphRAG to overcome challenges in generating regulation-compliant PFDs and PIDs for industrial processes. The framework demonstrates its capability to efficiently synthesize multimodal data from publicly available online sources, construct ontological knowledge graphs, and address ODQA tasks with high contextual accuracy. These advancements underscore the potential of AI-driven automation to streamline process design and accelerate the industrial application of emerging material innovations. Future work will focus on incorporating first-principles-based simulation tools to enhance the framework's precision and reliability, further bridging the gap between computational insights and industrial-scale implementation.

\bibliography{aaai25}

\begin{thebibliography}{12}
\providecommand{\natexlab}[1]{#1}

\bibitem[{Abuelsaad et~al.(2024)Abuelsaad, Akkil, Dey, Jagmohan, Vempaty, and
  Kokku}]{abuelsaad2024-agente}
Abuelsaad, T.; Akkil, D.; Dey, P.; Jagmohan, A.; Vempaty, A.; and Kokku, R.
  2024.
\newblock Agent-E: From Autonomous Web Navigation to Foundational Design
  Principles in Agentic Systems.
\newblock arXiv:2407.13032.

\bibitem[{Ansari et~al.(2024)Ansari, Watchorn, Brown, and
  Brown}]{ansari2024dziner}
Ansari, M.; Watchorn, J.; Brown, C.~E.; and Brown, J.~S. 2024.
\newblock dZiner: Rational Inverse Design of Materials with AI Agents.
\newblock \emph{arXiv preprint arXiv:2410.03963}.

\bibitem[{Anthropic(2023)}]{anthropic_contextual_retrieval}
Anthropic. 2023.
\newblock Introducing Contextual Retrieval.

\bibitem[{Halim et~al.(2022)Halim, Lakshmanan, Chen, Kumar, DasGupta, and
  Dehn}]{halim2022potential}
Halim, M.; Lakshmanan, V.; Chen, J.; Kumar, S.; DasGupta, S.; and Dehn, M.
  2022.
\newblock Potential Processes for Producing High-Purity Lithium Hydroxide: A
  Critical Review.
\newblock In \emph{Conference of Metallurgists}, 645--653. Springer.

\bibitem[{He et~al.(2024)He, Yao, Ma, Yu, Dai, Zhang, Lan, and
  Yu}]{he2024webvoyager}
He, H.; Yao, W.; Ma, K.; Yu, W.; Dai, Y.; Zhang, H.; Lan, Z.; and Yu, D. 2024.
\newblock WebVoyager: Building an End-to-End Web Agent with Large Multimodal
  Models.
\newblock \emph{arXiv preprint arXiv:2401.13919}.

\bibitem[{Jia, Zhang, and Fung(2024)}]{jia2024llmatdesign}
Jia, S.; Zhang, C.; and Fung, V. 2024.
\newblock LLMatDesign: Autonomous Materials Discovery with Large Language
  Models.
\newblock \emph{arXiv preprint arXiv:2406.13163}.

\bibitem[{Kim et~al.(2024)Kim, Choi, Kang, Lee, and Na}]{kim2024materials}
Kim, H.; Choi, H.; Kang, D.; Lee, W.~B.; and Na, J. 2024.
\newblock Materials discovery with extreme properties via reinforcement
  learning-guided combinatorial chemistry.
\newblock \emph{Chemical Science}.

\bibitem[{Lewis et~al.(2020)Lewis, Perez, Piktus, Petroni, Karpukhin, Goyal,
  K{\"u}ttler, Lewis, Yih, Rockt{\"a}schel et~al.}]{lewis2020retrieval}
Lewis, P.; Perez, E.; Piktus, A.; Petroni, F.; Karpukhin, V.; Goyal, N.;
  K{\"u}ttler, H.; Lewis, M.; Yih, W.-t.; Rockt{\"a}schel, T.; et~al. 2020.
\newblock Retrieval-augmented generation for knowledge-intensive nlp tasks.
\newblock \emph{Advances in Neural Information Processing Systems}, 33:
  9459--9474.

\bibitem[{Liu et~al.(2023)Liu, Jovanovic, Mallayya, Maddox, Wilson, Klemenz,
  Schoop, and Kim}]{liu2023materials}
Liu, Y.; Jovanovic, M.; Mallayya, K.; Maddox, W.~J.; Wilson, A.~G.; Klemenz,
  S.; Schoop, L.~M.; and Kim, E.-A. 2023.
\newblock Materials Expert-Artificial Intelligence for Materials Discovery.
\newblock \emph{arXiv preprint arXiv:2312.02796}.

\bibitem[{Putta et~al.(2024)Putta, Mills, Garg, Motwani, Finn, Garg, and
  Rafailov}]{putta2024agentqadvancedreasoning}
Putta, P.; Mills, E.; Garg, N.; Motwani, S.; Finn, C.; Garg, D.; and Rafailov,
  R. 2024.
\newblock Agent Q: Advanced Reasoning and Learning for Autonomous AI Agents.
\newblock arXiv:2408.07199.

\bibitem[{Radford et~al.(2021)Radford, Kim, Hallacy, Ramesh, Goh, Agarwal,
  Sastry, Askell, Mishkin, Clark et~al.}]{radford2021learning}
Radford, A.; Kim, J.~W.; Hallacy, C.; Ramesh, A.; Goh, G.; Agarwal, S.; Sastry,
  G.; Askell, A.; Mishkin, P.; Clark, J.; et~al. 2021.
\newblock Learning transferable visual models from natural language
  supervision.
\newblock In \emph{International conference on machine learning}, 8748--8763.
  PMLR.

\bibitem[{Sotelo et~al.(2017)Sotelo, Favela-Contreras, Sotelo, Jim{\'e}nez, and
  Gallegos-Canales}]{sotelo2017design}
Sotelo, D.; Favela-Contreras, A.; Sotelo, C.; Jim{\'e}nez, G.; and
  Gallegos-Canales, L. 2017.
\newblock Design and implementation of a control structure for quality products
  in a crude oil atmospheric distillation column.
\newblock \emph{ISA transactions}, 71: 573--584.

\end{thebibliography}

\clearpage
\newpage

\section{Technical Appendix}

\subsection{Additional Experiments}
\begin{figure*}[!ht]
    \centering

    \begin{subfigure}[a]{0.8\linewidth}
        \centering
        \resizebox{\linewidth}{!}{
            \hspace*{0mm}\includegraphics[keepaspectratio,trim=0.0cm 0.025cm 0cm 0.025cm,clip]{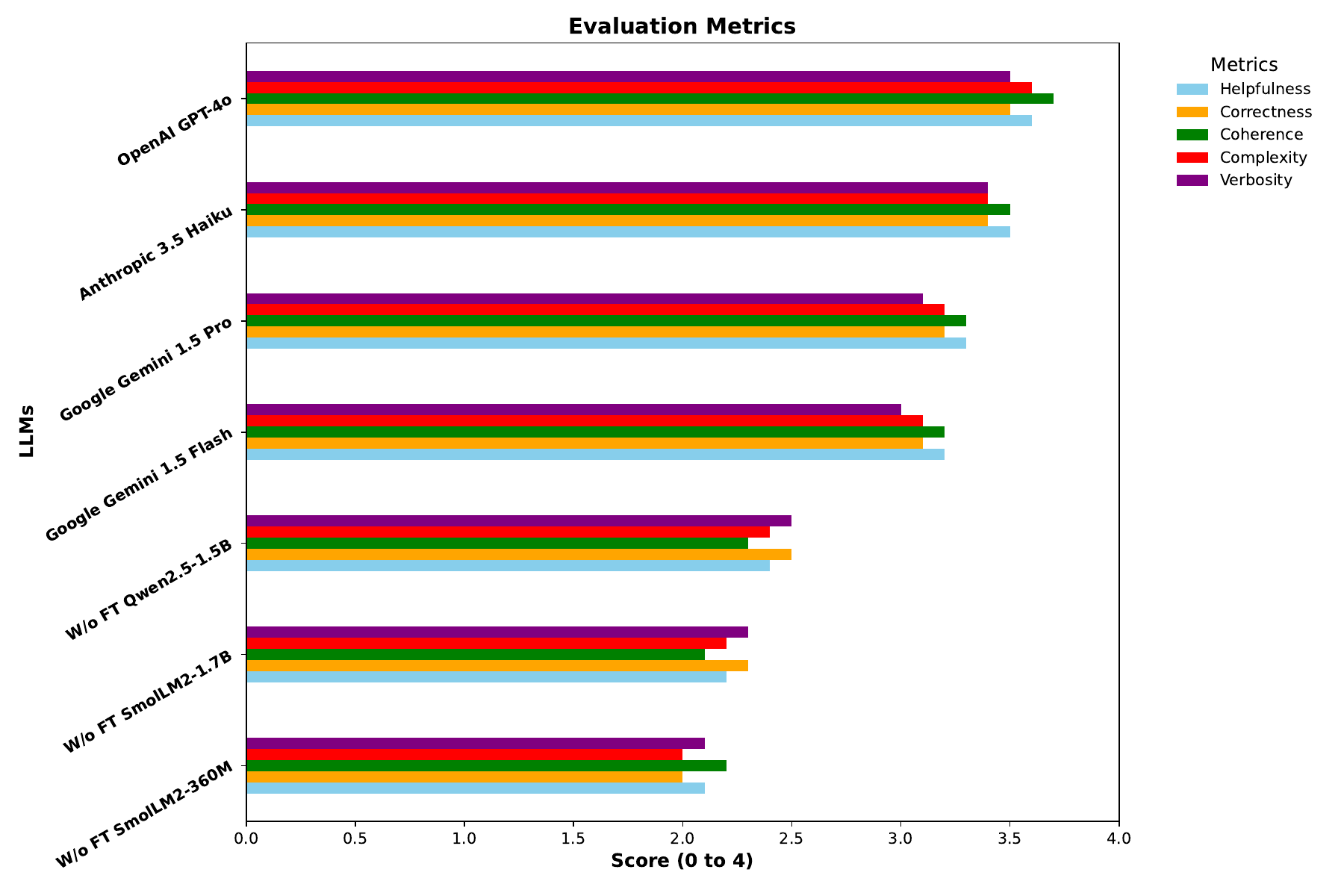} 
        }
        \caption{The figure illustrates the performance comparison between Graph RAG with pre-trained LLMs (W/o FT) and advanced closed-source LLMs (without Graph RAG) across various evaluation metrics.}
        \label{fig:figure12}
    \end{subfigure}

    \begin{subfigure}[b]{0.8\linewidth}
        \centering
        \resizebox{\linewidth}{!}{
            \hspace*{0mm}\includegraphics[keepaspectratio,trim=0.0cm 0.025cm 0cm 0.025cm,clip]{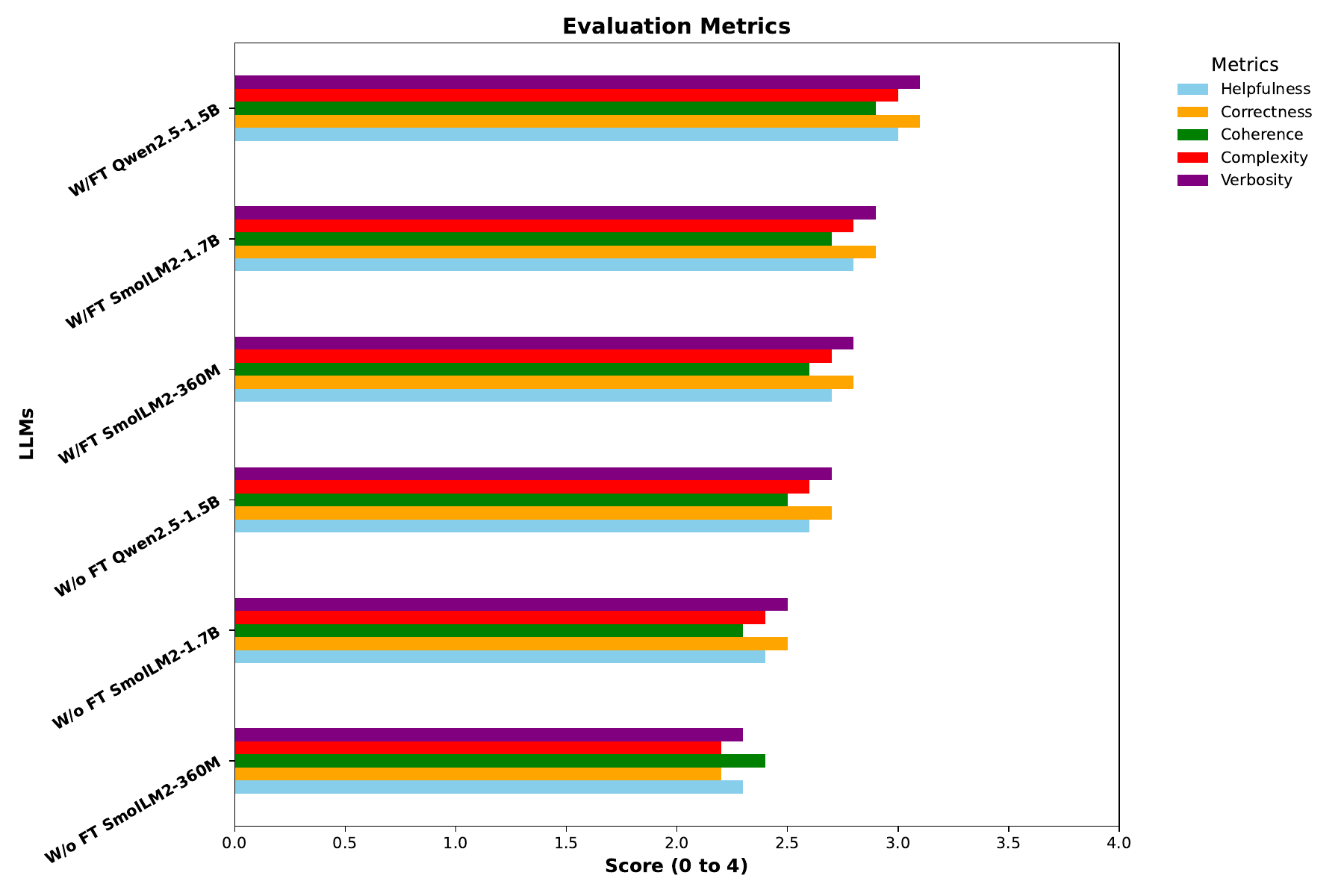}  
        }
        \caption{The figure shows the impact of Graph RAG with pre-trained LLMs (without fine-tuning) compared to fine-tuned LLMs (without Graph RAG).}
        \label{fig:figure13}
    \end{subfigure}

    \caption{The figures compare fine-tuned LLMs without Graph RAG to Graph RAG with pre-trained LLMs (without fine-tuning), evaluated across metrics such as helpfulness, correctness, coherence, complexity, and verbosity, scored on a scale from 0 to 4, similar to a Likert-style evaluation.}
    \vspace{-5mm}
\end{figure*}

We evaluate the impact of Graph RAG integrated with a pre-trained LLM (without fine-tuning, W/o FT) on generating PFDs and PIDs for the secondary evaluation subset of chemicals. Figure \ref{fig:figure12} presents the evaluation metrics for this setup. To establish a baseline, we assess proprietary, closed-source models such as GPT-4o and Google Gemini-1.5 Pro by directly evaluating their performance on unseen chemicals from the secondary evaluation subset without applying Graph RAG. The metrics—Helpfulness, Correctness, Coherence, Complexity, and Verbosity—are computed using the NVIDIA Nemotron-4-340B-Reward model, which evaluates responses based on usefulness, factual accuracy, clarity, intellectual depth, and appropriate detail. Scores are provided on a scale from 0 (lowest performance) to 4 (highest performance). Proprietary models like GPT-4o and Google Gemini-1.5 Pro consistently achieve higher scores across all metrics, demonstrating their advanced capabilities. In contrast, Graph RAG integrated with smaller models, such as SmolLM2-360M (W/o FT), scores relatively lower but still delivers acceptable performance in resource-constrained scenarios. These findings highlight that while state-of-the-art models set a high benchmark across all evaluation dimensions, smaller-scale models coupled with Graph RAG present a cost-efficient and practical alternative for many applications. We also compare Graph RAG with a pre-trained LLM (without fine-tuning) and fine-tuned LLMs (without Graph RAG) to evaluate the impact of structured retrieval on multi-hop reasoning and contextual retrieval for generating PFDs and PIDs in the secondary evaluation subset of chemicals. Figure \ref{fig:figure13} illustrates the performance of various Graph RAG configurations with pre-trained LLMs in comparison to fine-tuned LLMs (without Graph RAG). These configurations are assessed using evaluation metrics—Helpfulness, Correctness, Coherence, Complexity, and Verbosity—scored on a scale of 0 (lowest) to 4 (highest). The scores are determined by the NVIDIA Nemotron-4-340B-Reward model, which evaluates the quality of the generated text. Notably, fine-tuned models such as SmolLM2-360M, SmolLM2-1.7B, and Qwen2.5-1.5B achieve higher scores across all metrics compared to their pre-trained counterparts with Graph RAG, demonstrating the benefits of fine-tuning for tasks requiring multi-hop reasoning and contextual understanding. However, pre-trained models integrated with Graph RAG also exhibit competitive performance, underscoring their potential as cost-effective solutions for resource-constrained applications. Figure \ref{fig:figure18} compares the performance of Graph RAG with various pre-trained language models (W/o FT) and fine-tuned LLMs (without Graph RAG) on ODQA tasks using BLEU, ROUGE-N, and ROUGE-L metrics. The results indicate that fine-tuning and increased model size significantly enhance performance, with fine-tuned models (without Graph RAG) consistently outperforming their pre-trained counterparts integrated with Graph RAG. Notably, Graph RAG integrated with pre-trained models achieves performance levels close to those of the best-performing fine-tuned large models, demonstrating its ability to efficiently improve semantic and contextual alignment. This analysis underscores the importance of fine-tuning, model scaling, and the integration of Graph RAG in enhancing accuracy and coherence in complex language understanding tasks. In addition, Figure \ref{fig:figure19} presents evaluation metrics—Helpfulness, Correctness, Coherence, Complexity, and Verbosity—scored by the Nemotron-4-340B-Reward model across different configurations, showing that fine-tuned models (without Graph RAG) consistently outperform their pre-trained counterparts integrated with Graph RAG (W/o FT). Figure \ref{fig:figure20} demonstrates that Graph RAG with pre-trained LLMs (W/o FT) achieves significant improvements over traditional RAG with pre-trained LLMs (W/o FT) by leveraging structured knowledge graphs to enhance reasoning and understanding for the PFD and PID generation tasks of unknown chemicals.

\clearpage
\newpage

\begin{figure}[!ht]
    \centering
      \begin{subfigure}[a]{0.785\linewidth}
        \centering
        \resizebox{\linewidth}{!}{
            \hspace*{0mm}\includegraphics[keepaspectratio,trim=0.0cm 0.025cm 0cm 0.025cm,clip]{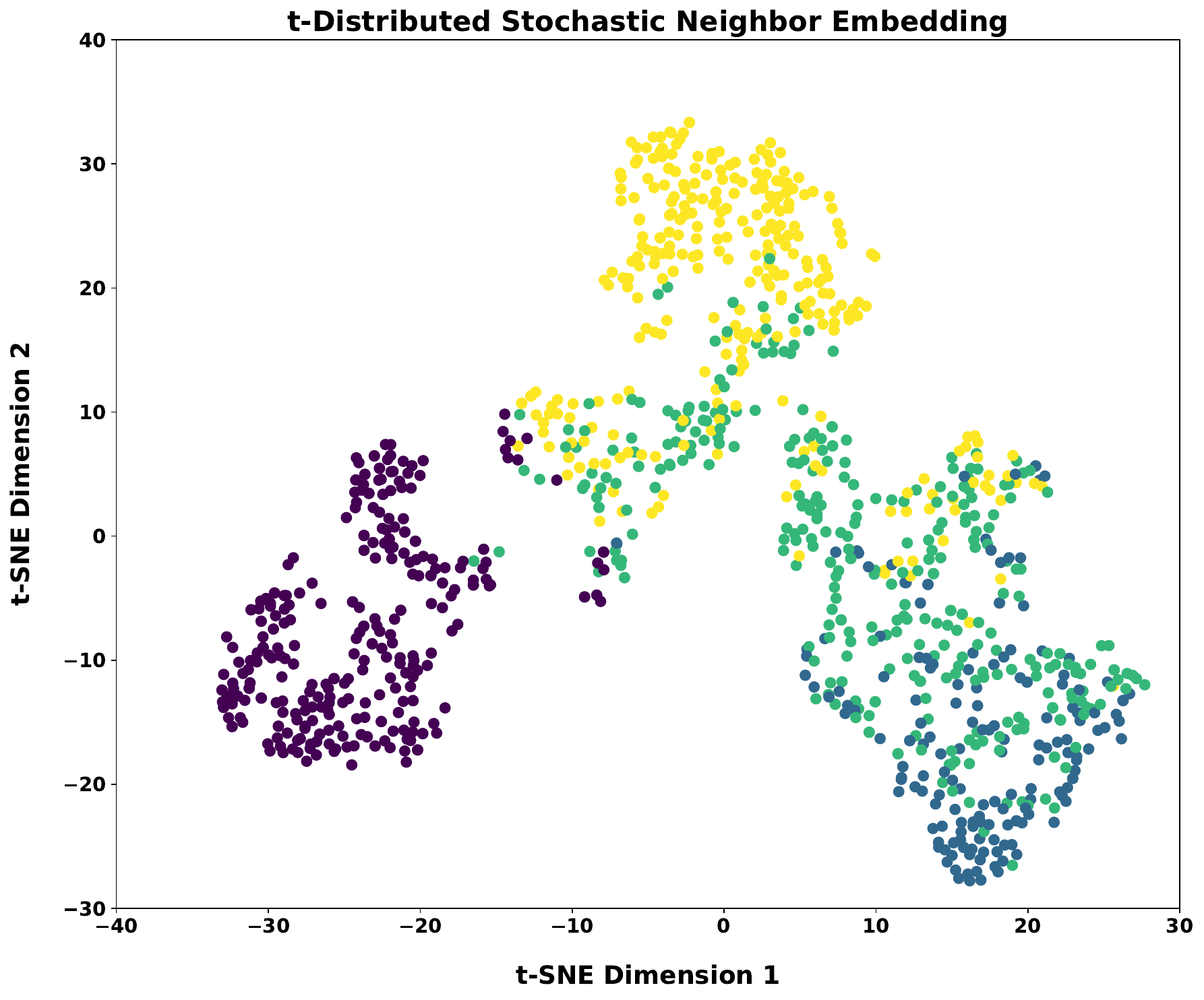} 
        }
        \caption{t-SNE plot: Visualization of PFD and PID data embeddings, showing non-linear clustering based on shared semantic features.}
        \label{fig:figure14}
    \end{subfigure}
    \begin{subfigure}[b]{0.785\linewidth}
        \centering
        \resizebox{\linewidth}{!}{
            \hspace*{0mm}\includegraphics[keepaspectratio,trim=0.0cm 0.025cm 0cm 0.025cm,clip]{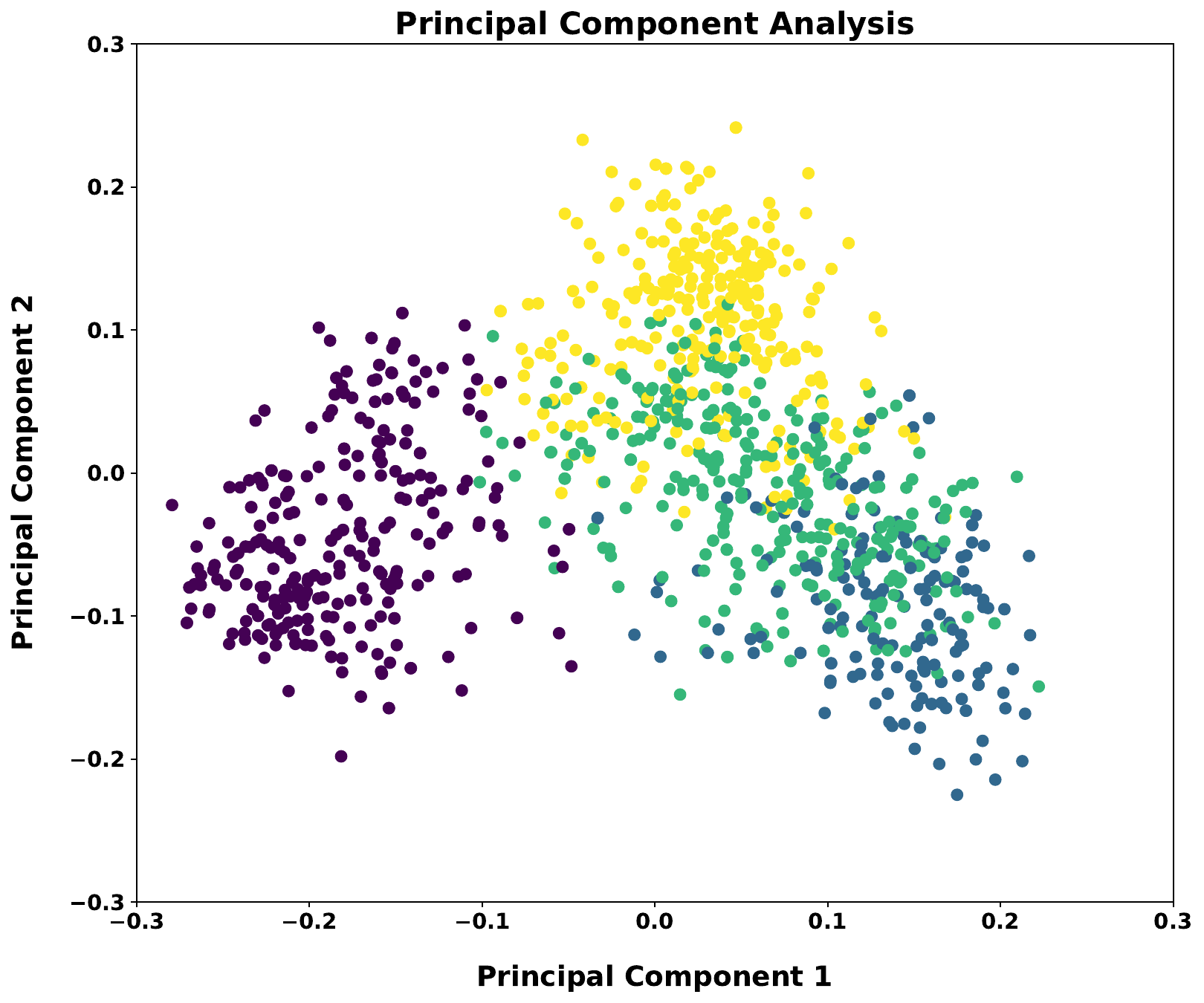}  
        }
        \caption{PCA plot: Principal component analysis of PFD and PID data embeddings, highlighting variance distribution and linear patterns.}
        \label{fig:figure15}
    \end{subfigure}
    \vspace{-2mm}
    \caption{The figure shows the t-SNE and PCA visualizations of Anthropic Haiku-generated PFD and PID data embeddings, highlighting non-linear clustering and variance distribution.}
    \vspace{-5mm}
\end{figure}

\begin{figure}[!ht]
    \centering
    \begin{subfigure}[a]{0.785\linewidth}
        \centering
        \resizebox{\linewidth}{!}{
            \hspace*{0mm}\includegraphics[keepaspectratio,trim=0.0cm 0.025cm 0cm 0.025cm,clip]{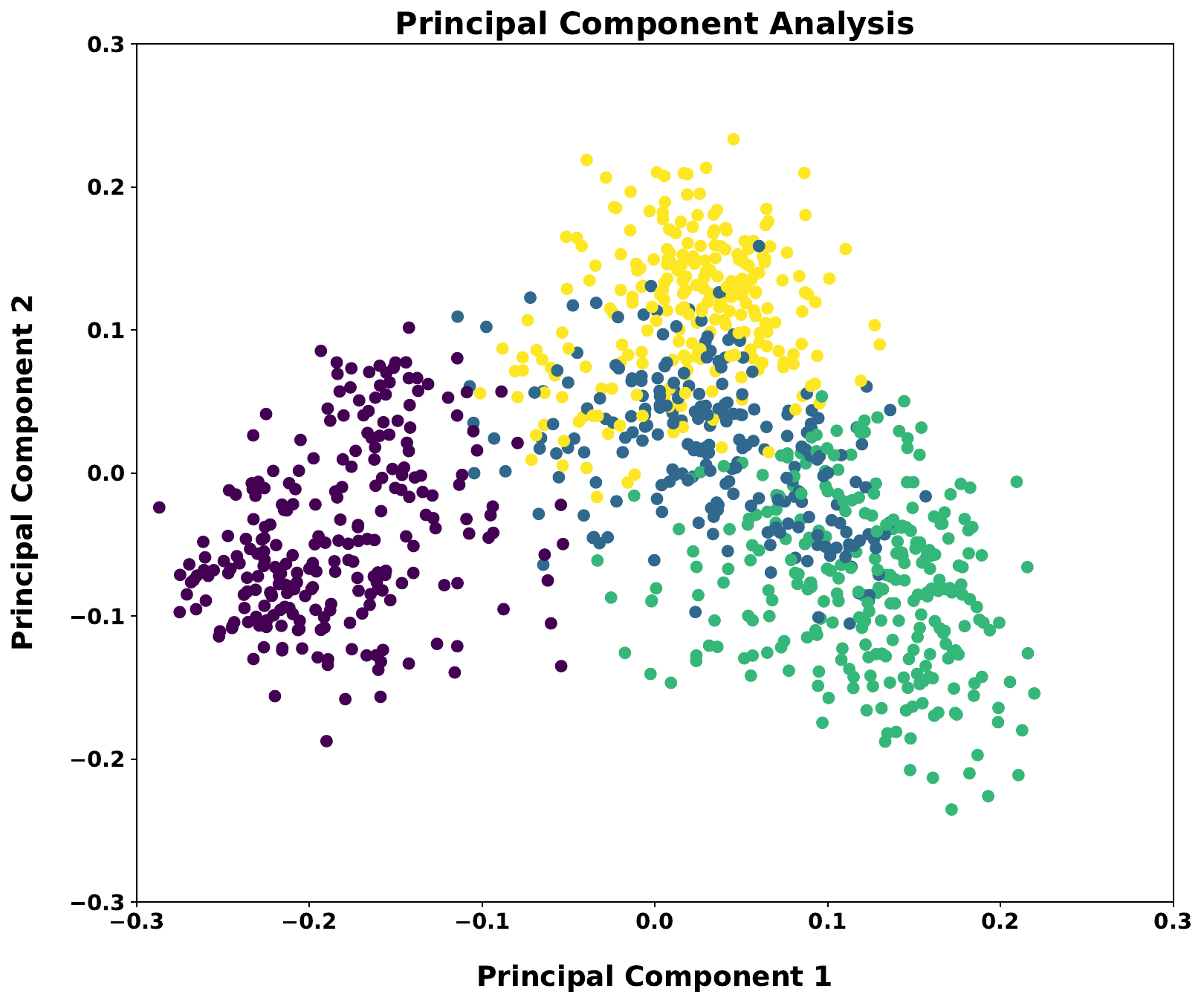} 
        }
        \caption{PCA plot (GPT-4o): Highlights the variance distribution and linear patterns in PFD and PID embeddings.}
        \label{fig:figure16}
    \end{subfigure}
    \begin{subfigure}[b]{0.785\linewidth}
        \centering
        \resizebox{\linewidth}{!}{
            \hspace*{0mm}\includegraphics[keepaspectratio,trim=0.0cm 0.025cm 0cm 0.025cm,clip]{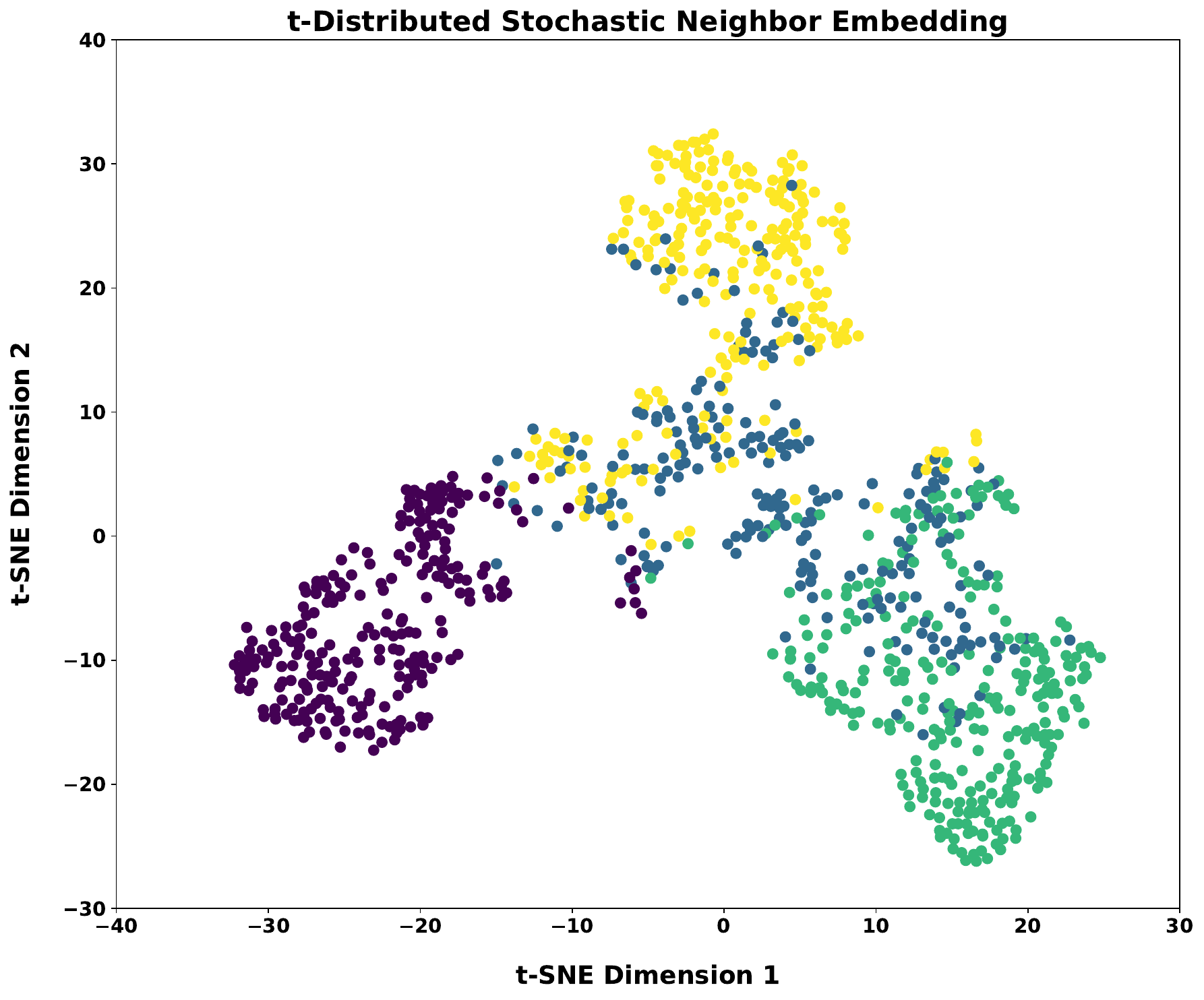}  
        }
        \caption{t-SNE plot (GPT-4o): Reveals clustering and non-linear relationships in PFD and PID embeddings.}
        \label{fig:figure17}
    \end{subfigure}
    \vspace{-2mm}
    \caption{The figure shows the visualizations of GPT-4o-generated embeddings for PFD and PID data, using PCA to show variance patterns and t-SNE to highlight clustering and non-linear structures.}
    \vspace{-5mm}
\end{figure}

\begin{figure}[!ht]
        \centering
        \resizebox{1.0\linewidth}{!}{
            \hspace*{-30mmmm}\includegraphics[keepaspectratio,trim=0.0cm 0.025cm 0cm 0.025cm,clip]{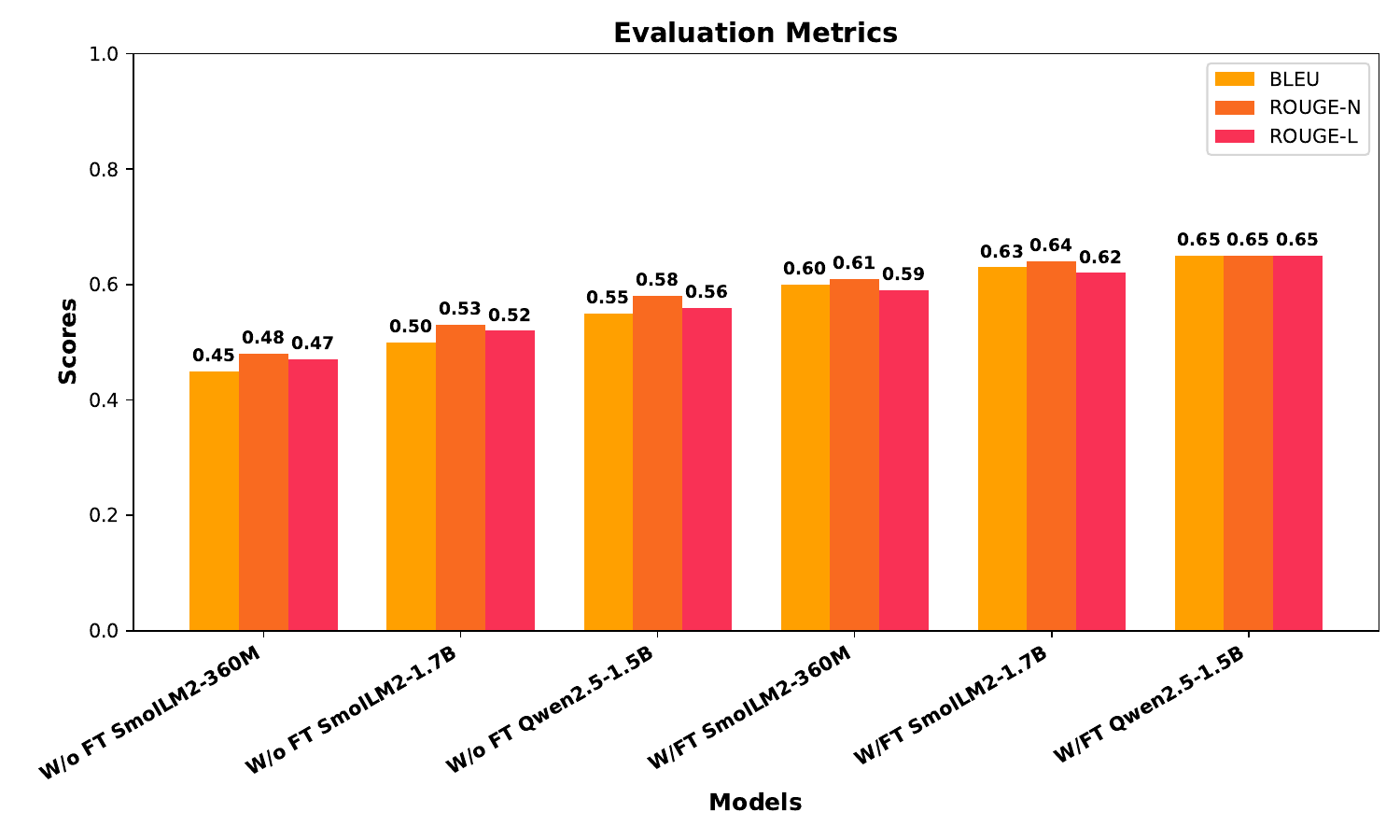}
        }
        \vspace{-8mm}
        \caption{The figure compares BLEU, ROUGE-N, and ROUGE-L metrics for fine-tuned models without Graph RAG and pre-trained models with Graph RAG(W/o FT), highlighting the effects of fine-tuning and model size.}
    \label{fig:figure18}
    \vspace{-3mm}
\end{figure}

\begin{figure}[!ht]
        \centering
        \resizebox{1.0\linewidth}{!}{
            \includegraphics[keepaspectratio,trim=0.0cm 0.025cm 0cm 0.025cm,clip]{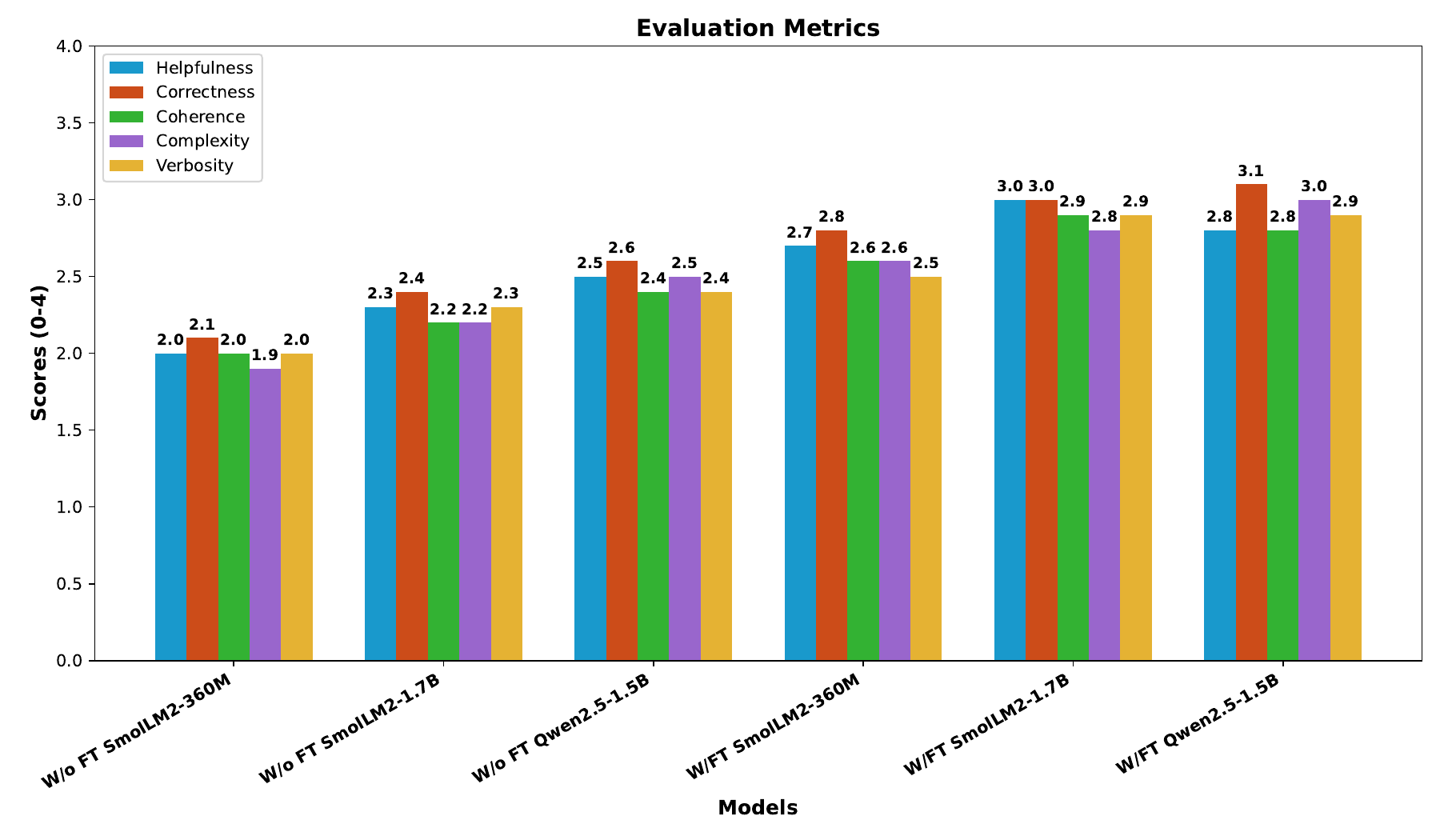}   
        }
        \vspace{-7mm}
        \caption{The figure shows the evaluation metrics (Helpfulness, Correctness, Coherence, Complexity, Verbosity) across different model configurations. Fine-tuned models (without Graph RAG) consistently achieve higher scores than Graph RAG with pre-trained models, based on results from the Secondary Evaluation Dataset. Scores are rated on a 0-4 scale.}
    \label{fig:figure19}
    \vspace{-3mm}
\end{figure}

\begin{figure}[!ht]
        \centering
        \resizebox{1.0\linewidth}{!}{
            \includegraphics[keepaspectratio,trim=0.0cm 0.025cm 0cm 0.025cm,clip]{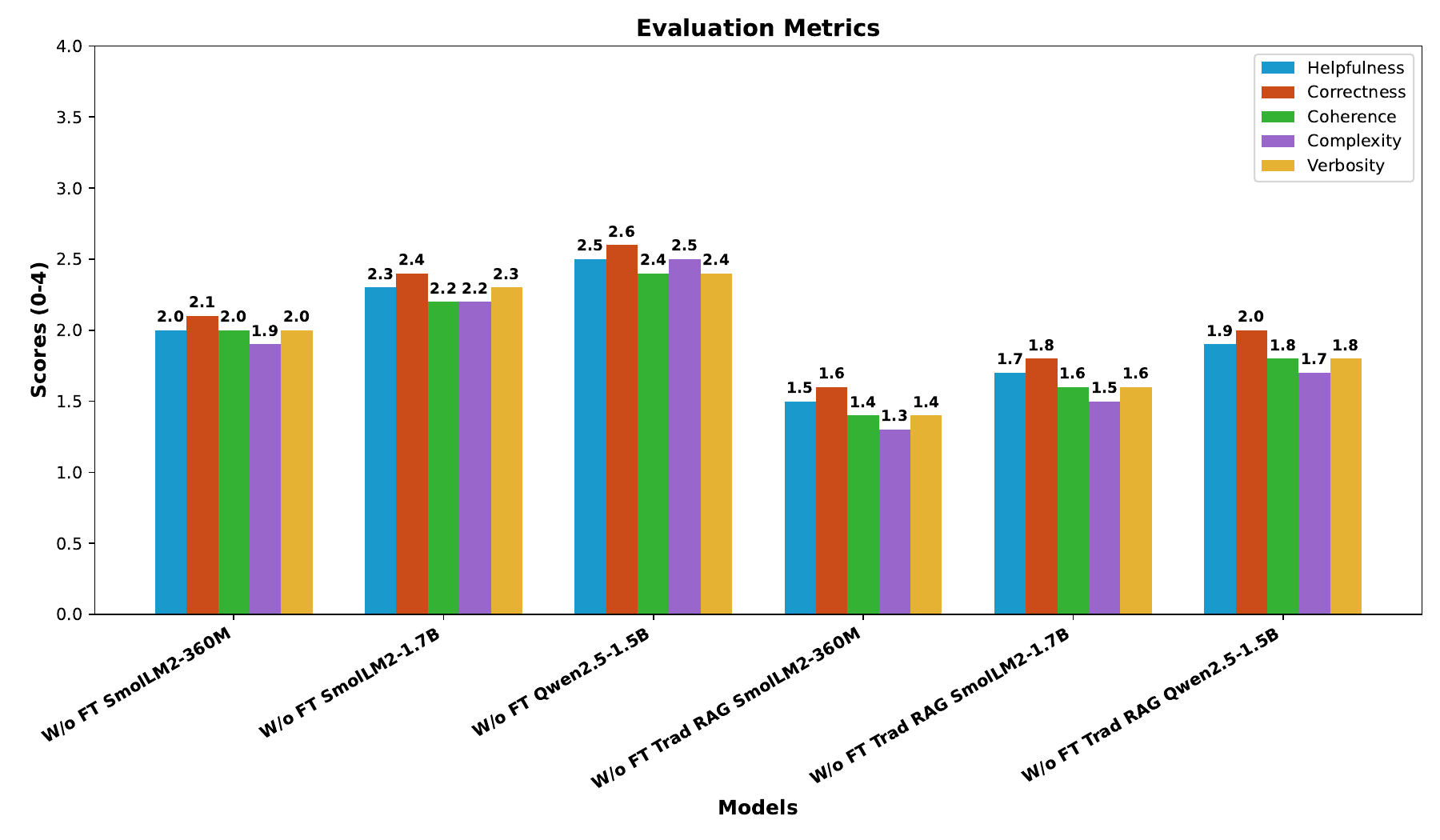}   
        }
        \vspace{-7mm}
        \caption{The figure shows a comparison of Graph RAG and traditional RAG with pre-trained LLMs on evaluation metrics for generating PFDs and PIDs for unknown chemicals from the secondary evaluation dataset. Graph RAG with pre-trained LLMs demonstrates superior performance as scored by the Nemotron-4-340B-Reward model.}
    \label{fig:figure20}
    \vspace{-3mm}
\end{figure}

Figures \ref{fig:figure14}-\ref{fig:figure15} present visualizations of chemical process data embeddings (PFDs and PIDs generated by Anthropic Haiku on the primary dataset) using two distinct dimensionality reduction techniques. The t-SNE plot (Figure \ref{fig:figure14}) captures non-linear relationships within the embeddings, revealing distinct clusters that highlight semantic groupings of PFD and PID data based on shared features. In contrast, the PCA plot (Figure \ref{fig:figure15}) illustrates the variance distribution along principal components, offering a linear perspective on the dataset and showcasing broader patterns with less pronounced clustering. Figures \ref{fig:figure16}-\ref{fig:figure17} depict data embeddings from PFD and PID knowledge generated by OpenAI GPT-4o on the primary Dataset (Knowledge Base), reduced to two dimensions using t-SNE and PCA techniques. The t-SNE visualization (Figure \ref{fig:figure16}) highlights clusters, uncovering intricate non-linear relationships within the dataset. The PCA plot (Figure \ref{fig:figure17}) illustrates the variance distribution, providing a linear perspective on broader patterns. Together, these visualizations offer complementary insights into the structure and organization of the processed information.

\vspace{-2mm}

\vspace{-3mm}

\begin{algorithm}[H]
\footnotesize
\caption{Autonomous Agentic Web Navigation Framework for Knowledge Generation}
\label{alg:autonomous_navigation_framework}
\begin{algorithmic}[1]
\Require Query $Q$ related to PFDs or PIDs on primary dataset
\Ensure Generated knowledge $A^*$ optimized through iterative refinement

\State \textbf{Initialize} Meta-Agent with task $Q$
\State \textbf{Decompose} $Q$ into subtasks $\{q_1, q_2, \dots, q_n\}$

\For{each subtask $q_i$}
    \For{each sub-agent $t_j$}
        \State Compute similarity:
        \[
        \text{sim}(v(q_i), v(d_j)) = \frac{v(q_i) \cdot v(d_j)}{\|v(q_i)\| \|v(d_j)\|}
        \]
    \EndFor
    \State \textbf{Select} sub-agent:
    \[
    t_j = \arg\max_{t_j} \text{sim}(v(q_i), v(d_j))
    \]
    \State \textbf{Invoke} $t_j$ with parameters $p_1$, $p_2$, $p_3$
    \State \textbf{Receive} result $R_{q_i}$ from $t_j$
\EndFor

\State \textbf{Manage} dependencies using DAG $\mathcal{G} = (\mathcal{V}, \mathcal{E})$
\State \textbf{Execute} independent subtasks in parallel

\For{each result $R_{q_i}$}
    \If{$t_j$ is the image agent}
        \State Retrieve images $I = \{i_1, \dots, i_k\}$
        \State Compute embeddings $e_I$ using CLIP
        \State Compute similarity:
        \[
        \text{sim}(e_I, e_{q_i}) = \frac{e_I \cdot e_{q_i}}{\|e_I\| \|e_{q_i}\|}
        \]
        \State Select relevant images $I_J \subset I$
        \State Generate summary $D_I$ using LLM
    \ElsIf{$t_j$ is the scholarly article agent}
        \State Retrieve articles $A = \{a_1, a_2, \dots, a_m\}$
        \State Compute embeddings $e_A$
        \State Compute similarity $\text{sim}(e_A, e_{q_i})$
        \State Select relevant articles $A_J \subset A$
        \State Generate summary $D_A$ using LLM
    \ElsIf{$t_j$ is the patent agent}
        \State Retrieve patents $P = \{p_1, p_2, \dots, p_l\}$
        \State Compute embeddings $e_P$
        \State Compute similarity $\text{sim}(e_P, e_{q_i})$
        \State Select relevant patents $P_J \subset P$
        \State Generate summary $D_P$ using LLM
    \ElsIf{$t_j$ is the wiki agent}
        \State Retrieve pages $W = \{w_1, w_2, \dots, w_p\}$
        \State Compute embeddings $e_W$
        \State Compute similarity $\text{sim}(e_W, e_{q_i})$
        \State Select relevant pages $W_J \subset W$
        \State Generate summary $D_W$ using LLM
    \ElsIf{$t_j$ is the web insights agent}
        \State Retrieve insights $G = \{g_1, g_2, \dots, g_q\}$
        \State Compute embeddings $e_G$
        \State Compute similarity $\text{sim}(e_G, e_{q_i})$
        \State Select relevant insights $G_J \subset G$
        \State Generate summary $D_G$ using LLM
    \EndIf
\EndFor
\State \textbf{Aggregate} outputs into response:
\[
A = \text{MetaAgent}_{\text{LLM}}(\text{Synthesize}(D_I, D_A, D_P, D_W, D_G))
\]
\State \textbf{Initialize} $i = 0$, $A_0 = A$
\end{algorithmic}
\end{algorithm}

\begin{algorithm}[H]
\footnotesize
\begin{algorithmic}[1]
\setcounter{ALG@line}{47} 

\Repeat
    \State \textbf{Generate} feedback $F_i$ from experts and benchmark models
    \State \textbf{Update} output:
    \[
    A_{i+1} = \text{MetaAgent}_{\text{LLM}}(Q, A_i, F_i)
    \]
    \State $i \gets i + 1$
\Until{$A_i$ meets quality standards or $i \geq N_{\text{max}}$}

\State \textbf{Set} optimized generated knowledge $A^* = A_i$
\State \textbf{Store} $A^*$ in documents $D$

\end{algorithmic}
\end{algorithm}

\vspace{-4mm}
\begin{algorithm}[H]
\footnotesize
\caption{Knowledge Graph Construction and Utilization in ODQA}
\label{alg:knowledge_graph_ODQA}
\begin{algorithmic}[1]
\Require Input documents $D = \{D^{(1)}, D^{(2)}, \dots, D^{(M)}\}$, ODQA query $Q$, window size $w$, stride $s$, thresholds $\tau_{\text{sim}}$ and $\tau_{\text{str}}$
\Ensure Optimal answer $A_q$ for the ODQA task

\State \textbf{Initialize} an empty knowledge graph $\mathcal{G} = (\mathcal{V}, \mathcal{E})$

\For{each document $D^{(m)} \in D$}
    \State Divide $D^{(m)}$ into chunks $C_i^{(m)}$ using window size $w$ and stride $s$:
    \[
    C_i^{(m)} = D^{(m)}\left[(i - 1) \cdot s : (i - 1) \cdot s + w\right]
    \]
    \State Generate contextual summary $\text{ctx}_i^{(m)}$ for each chunk using a language model
    \State Enrich chunk by concatenating context:
    \[
    C_i^{\prime (m)} = \text{ctx}_i^{(m)} \oplus C_i^{(m)}
    \]
    \State Perform entity extraction to identify entities:
    \[
    E_i^{(m)} = \{e_j^{(m, i)} : j = 1, 2, \dots, K^{(m, i)}\}
    \]
    \State Perform relation extraction to identify relationships:
    \[
    R_i^{(m)} = \{(e_j^{(m, i)}, r_{j,k}^{(m, i)}, e_k^{(m, i)})\}
    \]
    \State Add entities $E_i^{(m)}$ and relations $R_i^{(m)}$ to $\mathcal{G}$
\EndFor

\State \textbf{Deduplicate entities:}
\For{each pair $(e_j^{(m, i)}, e_{j'}^{(m', i')})$ in $\mathcal{G}$}
    \State Compute semantic similarity:
    {\small
    \[
    \text{sim}(\mathbf{v}_{j}^{(m, i)}, \mathbf{v}_{j'}^{(m', i')}) = \frac{\mathbf{v}_{j}^{(m, i)} \cdot \mathbf{v}_{j'}^{(m', i')}}{\|\mathbf{v}_{j}^{(m, i)}\| \|\mathbf{v}_{j'}^{(m', i')}\|}
    \]}
    \State Compute string similarity: $  \text{str\_sim}(e_j^{(m, i)}, e_{j'}^{(m', i')}) $
    {\small
    \[
    \text{str\_sim}(e_j^{(m, i)}, e_{j'}^{(m', i')}) = 1 - \frac{d_{\text{lev}}(e_j^{(m, i)}, e_{j'}^{(m', i')})}{\max(|e_j^{(m, i)}|, |e_{j'}^{(m', i')}|)}
    \]}
    \If{$\text{sim}(\mathbf{v}_{j}^{(m, i)}, \mathbf{v}_{j'}^{(m', i')}) \geq \tau_{\text{sim}}$ and $\text{str\_sim}(e_j^{(m, i)}, e_{j'}^{(m', i')}) \geq \tau_{\text{str}}$}
        \State Merge entities $e_j^{(m, i)}$ and $e_{j'}^{(m', i')}$
    \EndIf
\EndFor

\State \textbf{Detect communities in $\mathcal{G}$ using the Leiden algorithm:}
\State Partition graph into communities $\{\mathcal{C}_1, \mathcal{C}_2, \dots, \mathcal{C}_L\}$ by maximizing modularity:
\end{algorithmic}
\normalsize
\end{algorithm}

\begin{algorithm}[H]
\footnotesize
\begin{algorithmic}[1]
\setcounter{ALG@line}{19} 
\State 
\[
Q_{\text{Mod}} = \frac{1}{2m} \sum_{i,j} \left[ A_{ij} - \frac{k_i k_j}{2m} \right] \delta(c_i, c_j)
\]
\State \textbf{Generate community summaries:}
\For{each community $\mathcal{C}_i$}
    \State Condense relationship paths $R_i$ into summaries $S_i$ using a language model:
    \[
    S_i = \text{LLM}(R_i) = \arg\max_{S} P(S \mid R_i)
    \]
    \State Convert summaries $S_i$ into vector embeddings $\mathbf{v}(S_i)$
\EndFor

\State \textbf{Retrieve query-specific subgraph:}
\State Rank communities $\{\mathcal{C}_1, \mathcal{C}_2, \dots, \mathcal{C}_L\}$ by similarity between $Q$ and community summaries:
\[
d(Q, \mathcal{C}_i) = \frac{\langle \mathbf{v}(Q), \mathbf{v}(S_i) \rangle}{\|\mathbf{v}(Q)\| \cdot \|\mathbf{v}(S_i)\|}
\]
\State Select top-$K$ communities and construct subgraph:
\[
\mathcal{V}_Q = \bigcup_{i=1}^K \mathcal{V}_{\mathcal{C}_i}, \quad \mathcal{E}_Q = \bigcup_{i=1}^K \mathcal{E}_{\mathcal{C}_i}
\]

\State Extract paths $P$ from $G_Q = (\mathcal{V}_Q, \mathcal{E}_Q)$ for reasoning
\State Use language model to generate answer $A_q$:
\[
A_q = \text{LLM}(Q, P) = \arg\max_{A} \, P(A \mid Q, P)
\]

\Return Optimal answer $A_q$

\end{algorithmic}
\end{algorithm}

\section{Additional details}
The hyperparameters for constructing knowledge graphs are optimized to enhance core processes, including chunking, triple extraction, similarity evaluation, and storage. Chunking parameters, such as window size (\(w\)) and stride (\(s\)), ensure that text is segmented into manageable and contextually continuous segments. The window size (\(w\)) is set to 1024 tokens, allowing for larger contextual segments to capture detailed information, while the stride (\(s\)) is set to 128 tokens, maintaining a 12.5\% overlap to balance contextual continuity and computational efficiency. For triple extraction, advanced LLMs, such as GPT-4o, identify entities and relationships, generating structured triples while adhering to a maximum per-chunk threshold (\(M\)) of 20 triples to balance graph complexity and efficiency. Similarity evaluation incorporates a high cosine similarity threshold (0.9) to ensure precise merging of semantically similar entities and applies a Levenshtein edit distance limit of 5 to resolve minor variations in entity names or labels effectively. Finally, the graph is stored in robust systems like Neo4j, capable of efficiently managing the increased data volume from larger window sizes. This parameterization ensures scalability, precision, and semantic consistency, leveraging state-of-the-art techniques for constructing and managing knowledge graphs. Optimization of the ranking and retrieval steps is crucial for handling large graphs efficiently, as these processes can be computationally expensive. To address this, precomputing summaries $(S_i)$ for frequently used communities can significantly reduce runtime by avoiding repetitive computations during queries. Additionally, indexing the embeddings $(v(S_i))$ of these summaries enables faster similarity computations, enhancing the overall scalability and framework performance.

\vspace{-3mm}

\vspace{-6mm}

\vspace{-4mm}
\begin{figure}[!ht]
        \centering
        \resizebox{1.0\linewidth}{!}{
            \includegraphics[keepaspectratio,trim=0.0cm 0cm 0cm 0.025cm,clip]{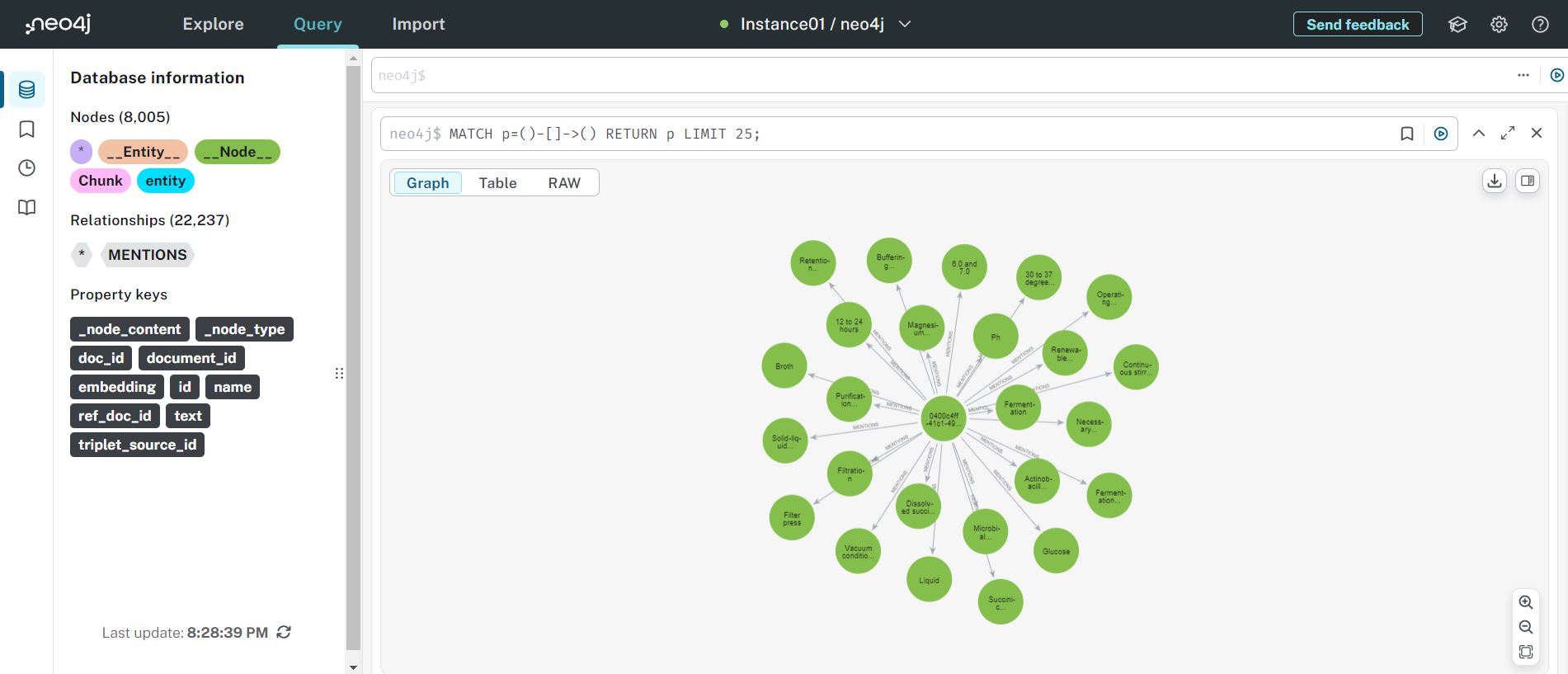}  
        }
        \vspace{-4mm}
        \caption{The figure shows a Neo4j graph database visualization comprising 8,005 nodes and 22,237 relationships. The nodes represent key entities or parent chunks from the knowledge graph. Entity nodes are connected to their respective parent chunk nodes and to other entities, indicating associations or references within the database.}
    \label{fig:figure21}
    \vspace{-3mm}
\end{figure}

\end{document}